\documentclass[manuscript,screen]{acmart}

\AtBeginDocument{%
  }

\setcopyright{acmlicensed}
\copyrightyear{2024}
\acmYear{2024}
\acmDOI{10.1145/3654660}

\acmConference[ACM Comput. Surv.]{}{}{}
\acmISBN{978-1-4503-XXXX-X/18/06}

\usepackage{csquotes}
\usepackage{adjustbox}
\usepackage{booktabs}
\usepackage{array,multirow,graphicx}
\usepackage[normalem]{ulem}
\usepackage{rotating}
\useunder{\uline}{\ul}{}
\usepackage{tikz}
\usepackage{calc}
\usepackage{csquotes}
\usepackage{natbib}
\usepackage{fontawesome}
\usepackage{pifont}
\usepackage{tcolorbox}

\usepackage{enumitem,kantlipsum}

\begin{document}

\fancyhead[LO,LE]{\textbf{\underline{Paper accepted at ACM Computing Surveys}}}

\title{Computational Politeness in Natural Language Processing: A Survey}

\author{Priyanshu Priya}
\authornote{Corresponding authors}
\email{priyanshu\_2021cs26@iitp.ac.in}
\affiliation{%
  \institution{Indian Institute of Technology Patna}
  \streetaddress{Bihta}
  \city{Patna}
  \state{Bihar}
  \country{India}
  \postcode{801106}
}

\author{Mauajama Firdaus}
\email{mauzama.03@gmail.com}
\affiliation{%
  \institution{University of Alberta}
  \city{Edmonton}
  \state{Alberta}
  \country{Canada}
}

\author{Asif Ekbal}
\authornotemark[1]
\email{asif@iitp.ac.in}

\affiliation{%
  \institution{Indian Institute of Technology Patna}
  \streetaddress{Bihta}
  \city{Patna}
  \state{Bihar}
  \country{India}
  \postcode{801106}
}

\renewcommand{\shortauthors}{Priya et al.}

\begin{abstract}
  Computational approach to politeness is the task of automatically predicting and/or generating politeness in text. This is a pivotal task for conversational analysis, given the ubiquity and challenges of politeness in interactions. The computational approach to politeness has witnessed great interest from the conversational analysis community. This article is a compilation of past works in computational politeness in natural language processing. We view four milestones in the research so far, \textit{viz.} supervised and weakly-supervised feature extraction to identify and induce politeness in a given text, incorporation of context beyond the target text, study of politeness across different social factors, and study the relationship between politeness and various socio-linguistic cues. In this article, we describe the datasets, approaches, trends, and issues in computational politeness research. We also discuss representative performance values and provide pointers to future works, as given in the prior works. In terms of resources to understand the state-of-the-art, this survey presents several valuable illustrations — most prominently, a table summarizing the past papers along different dimensions, such as the types of features, annotation techniques, and datasets used.
\end{abstract}




\received{4 August 2022}
\received[revised]{27 February 2024}
\received[accepted]{10 March 2024}

\maketitle

\section{Introduction}

The Free Dictionary\footnote{\url{https://www.thefreedictionary.com/}} defines the term polite as being \enquote{marked by or showing consideration for others and observance of accepted social usage}. The Oxford Learner's Dictionary\footnote{\url{https://www.oxfordlearnersdictionaries.com/}} defines the term \enquote{polite} as \enquote{having or showing good manners and respect for the feelings of others}. The term polite has been defined in multiple ways by the Merriam-Webster\footnote{\url{https://www.merriam-webster.com/}} dictionary like \enquote{showing or characterized by correct social usage}, \enquote{marked by an appearance of consideration, tact, deference, or courtesy} and \enquote{marked by a lack of roughness or crudities}. All the definitions of the term \textit{\textbf{polite}} encompass behavior that is socially correct and acceptable. According to \cite{lakoff1973logic,brown1987politeness}, politeness is a ubiquitous quality of human communication. In almost every situation, the speaker can be more or less polite to the listener, which might affect the speaker's social goals. Politeness is conveyed by a specific and extensive collection of linguistic markers that can alter the information content of an utterance. Occasionally, politeness is an act of commission such as saying \enquote{\textit{thank you}}, \enquote{\textit{you're welcome}} and \enquote{\textit{please}}, while the other times it is an act of omission, for instance, declining to be contradictory. Moreover, the mapping of politeness indicators is frequently context-specific (workplace vs. family vs. friends), speaker-specific (male vs. female), goal-specific (buyer vs. seller), or culture-specific. Although politeness often relies on a shared pool of linguistic markers for social coordination between the speaker and the listener, the importance and valence of each marker are often determined by the context, which poses a challenge for computational politeness studies.

The growth and development of Artificial Intelligence (AI) and Natural Language Processing (NLP) have led to a rise in the popularity of human-computer interaction. Consequently, the need to emulate human behavior has increased. Politeness is one of the most significant characteristics of human beings, and we often make decisions about when, where, and how to use politeness in communication to promote social interactions. Since politeness exemplifies the socially desirable behavior that helps regulate the essential aspects of social interaction, computational approaches to politeness are crucial for analyzing, simulating, and facilitating such interactions. 

The challenges and benefits of politeness to human-human and human-machine interactions have led to an interest in computational approaches to politeness as a research problem. The computational approach to politeness involves developing tools that automatically identify, evaluate, adapt, interpret, and generate linguistic cues conveying polite behavior in human-human and human-machine interactions. %Thus, the sentence \enquote{\textit{Can you help me with my work, please?}} should be perceived as polite by the computational model while the sentence \enquote{\textit{Oh! Let it be, please. It's not your cup of tea.}} should be interpreted as non-polite or, to say better, impolite. 
Developing a computational model of politeness and applying and evaluating it in different contexts is a challenging task due to the nuanced ways in which politeness may be expressed. 
\begin{figure*}
    \centering
    \begin{adjustbox}{max width=\linewidth}
    \includegraphics[scale=1.0]{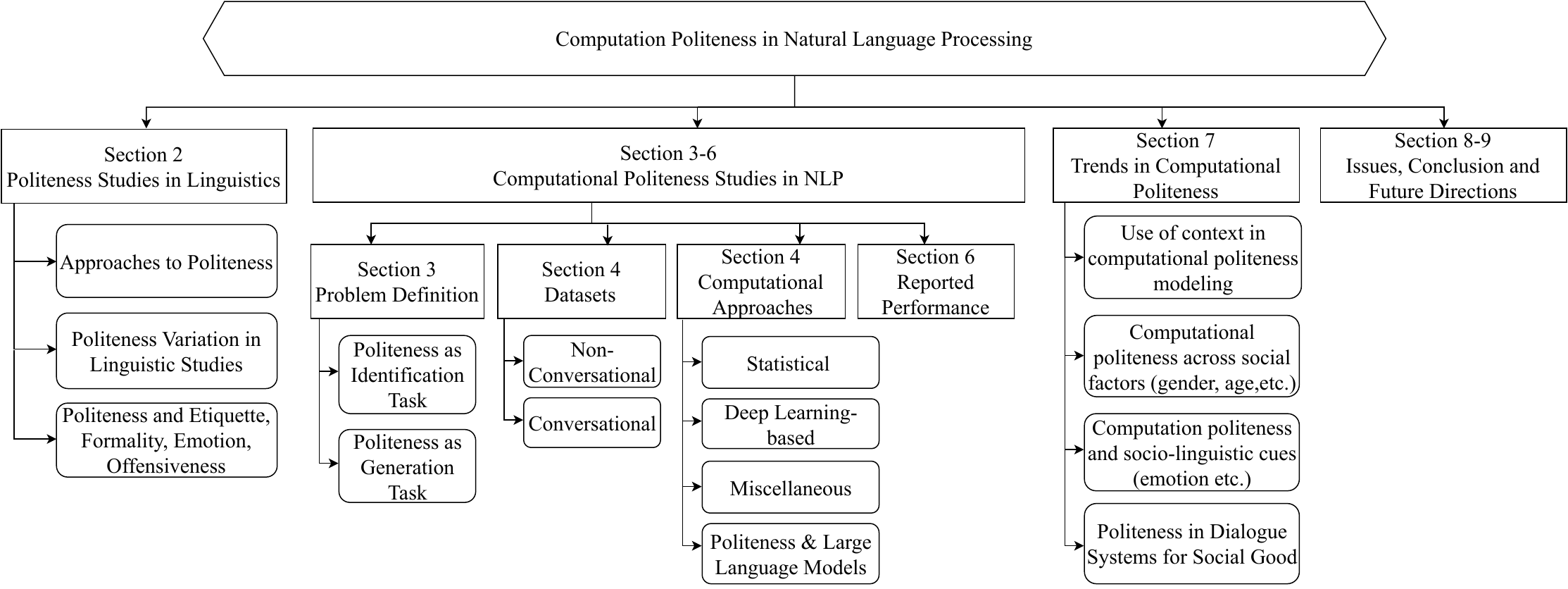}
    \end{adjustbox}
    \caption{Organization of the present survey}
    \label{organization}
    % \vspace{-6pt}
\end{figure*}
The initial work by \citet{danescu2013computational} on politeness leveraged foundational research in linguistic politeness \cite{lakoff1973logic,brown1987politeness} to discern linguistic markers of politeness in online requests. This was achieved through a computational framework based on Support Vector Machines (SVMs), which has since garnered significant attention within the Natural Language Processing (NLP) community.
% , who have utilized the foundational work in linguistic politeness \cite{lakoff1973logic,lakoff1977you,brown1987politeness} to identify linguistic aspects of politeness in online requests using a Support Vector Machine (SVM)-based computational framework has sparked the interest of the Natural Language Processing (NLP) community. 
The computational approach to politeness encompasses a variety of data types and methodologies. This combination has brought in some intriguing advancements in the field of computational politeness. 
% The primary objective of our study is to furnish a thorough overview of the previous work in the field of computational politeness. %We believe that this survey article will benefit the new researchers in grasping the state-of-the-art methods in this field.

As shown in Figure \ref{organization}, we organize the remainder of the article as follows: Section \ref{section2} describes the linguistic studies of politeness. To have a better understanding of the different aspects of previous work in computational politeness, this survey article examines politeness in six phases. Sections \ref{section3}, \ref{section4}, and \ref{section5} successively discuss the different problem definitions, datasets and approaches for computational politeness studies. In Section \ref{section6}, we present the reported performance values. Section \ref{section7} covers the underlying trends in computational politeness followed by the issues in Section \ref{section8}. Eventually, Section \ref{section9} concludes the article and makes recommendations for future research work. We have included tables and examples in this survey paper that may be used to gain a better understanding of computational politeness studies. The primary objective of this survey is to provide a comprehensive overview of previous work in the field of computational politeness, with insights into future research that may prove valuable to researchers interested in computational politeness and related fields.

%==================================================================================================%

\section{Politeness studies in Linguistics}
\label{section2}

Before discussing computational approaches to politeness, we outline a brief overview of linguistic studies about politeness in this section. Politeness is amongst the most crucial components of human communication, and it has the potential to determine the success or failure of any human interaction \cite{kumar2014developing}. Politeness studies have taken center stage in pragmatics research during the past few decades. These studies have mainly focused on applying various communicative tactics to preserve or promote social harmony. Politeness has been defined in different ways by different theoreticians. Some of the most notable definitions are as follows:
\begin{enumerate}[leftmargin=*,topsep=0pt]
\setlength{\itemsep}{0pt}
\setlength{\parskip}{0pt}
\setlength{\parsep}{0pt}
    \item \citet{leech1983principles} defines politeness as \enquote{maintain[ing] the social equilibrium and the friendly relations which enable us to assume that our interlocutors are being cooperative in the first place}.
    \item Another notable definition of politeness was proposed by \citet{brown1987politeness} %Brown and Levinson 
    as \enquote{politeness, like formal diplomatic protocol (for which it must surely be the model), presupposes that potential for aggression as it seeks to disarm it, and makes possible communication between potentially aggressive parties.}.
    \item  \citet{lakoff1989limits} defines politeness as \enquote{a means of minimizing confrontation in discourse - both the possibility of confrontation occurring at all, and the possibility that a confrontation will be perceived as threatening}.
    \item \citet{thomas2014meaning} summarizes the research agendas of linguists such as those mentioned above who are concerned with the study of pragmatic politeness as \enquote{All that is really being claimed is that people employ certain strategies (including the 50+ strategies described by Leech, Brown and Levinson, and others) for reasons of expediency – experience has taught us that particular strategies are likely to succeed in given circumstances, so we use them}.
    \item \citet{watts2003politeness} holds that politeness should be defined through a discursive approach. He believes that such struggle defines \enquote{the ways in which (im)polite behaviour is evaluated and commented on by lay members and not with ways in which social scientists lift the term (im)politeness out of the realm of everyday discourse and evaluate it to the status of a theoretical concept in what is frequently called Politeness Theory}.
    \item \citet{yule2020study} in his study asserts that individuals can be polite through \enquote{being tactful, modest and nice to other people}. 
\end{enumerate}

\subsection{Approaches to Politeness}
Several linguistic studies describe different approaches to politeness. Starting from the classic view of politeness that is based on traditional pragmatic theories, most notably Conversational Implicature \cite{grice1975logic} and Speech Act Theory \cite{searle1969speech} to post-modern view, politeness has been conceptualized as maxim-based, face-based, discursive, relational and frame-based approaches. 

\begin{figure}
    \centering
    %\begin{adjustbox}{max width=\linewidth}
    \includegraphics[height=8cm, width=13cm,scale=1.0]{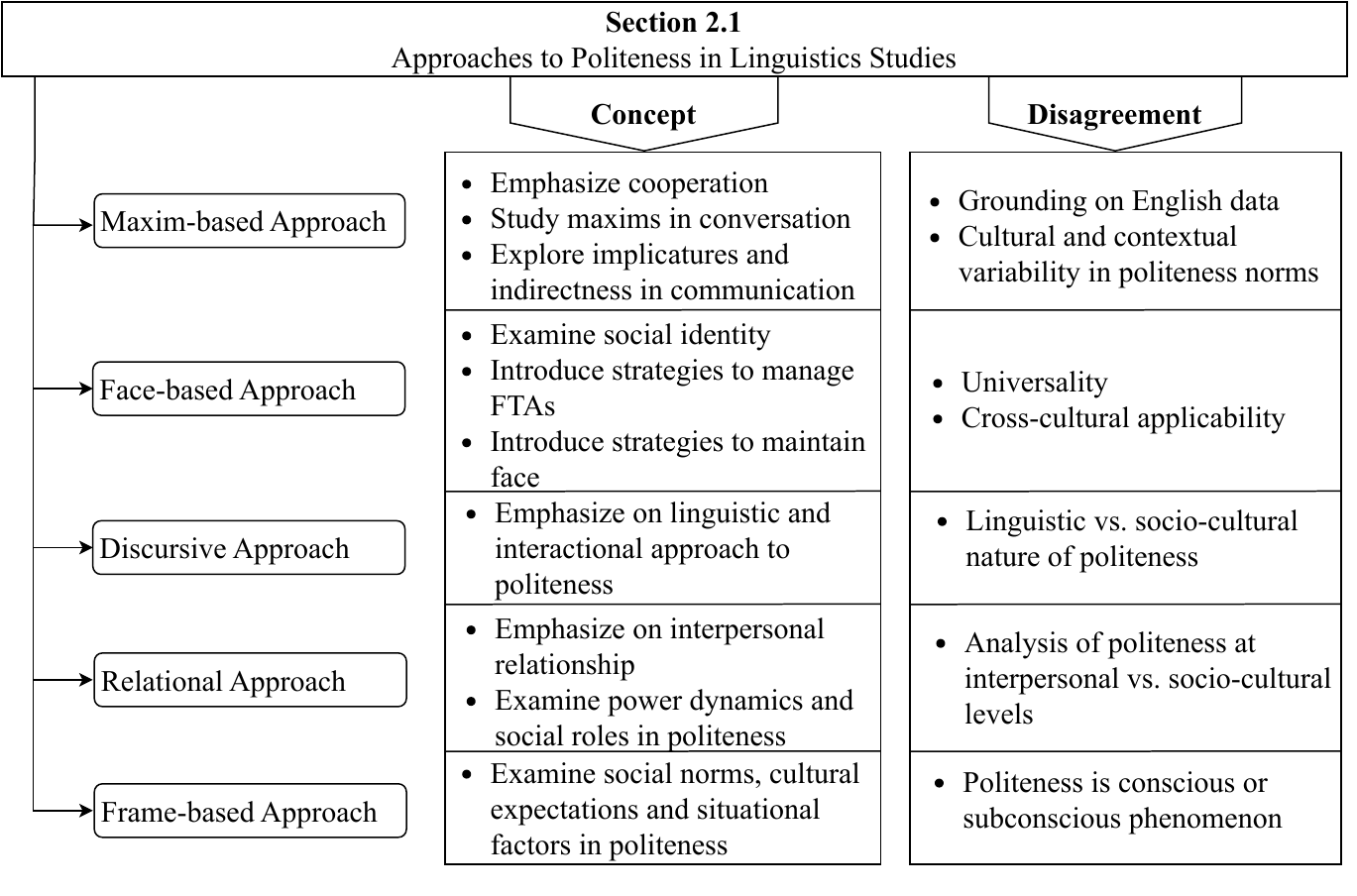}
    %\end{adjustbox}
    \caption{Overview of various approaches to politeness in linguistic studies}
    \label{pol_approaches}
    \vspace{-6pt}
\end{figure}

The overview of the different facets of politeness considered in these approaches and their significant disagreements/criticisms in the linguistic studies are depicted in Figure \ref{pol_approaches}. These different approaches to politeness offer distinct perspectives on how people manage politeness in communication. They differ in their emphasis on linguistic maxims, face, discourse, relationships, and social frames. %While researchers often disagree about which approach is most comprehensive or applicable in all contexts, each approach contributes valuable insights into understanding politeness in communication. The choice of approach may vary depending on the specific research question, context, and cultural considerations. }

% \begin{enumerate}[listparindent=0em]
% \setlength\itemsep{0em}
% \item 
\textbf{Maxim-based Approach to Politeness. }The language philosophers like Lakoff and Leech utilize maxims to describe politeness. Lakoff identifies two norms of pragmatic competence: (i) `be clear' and (ii) `be polite'. The former is formalized using \citet{grice1975logic} Cooperative Principle (CP), while the latter is formalized in terms of \citet{leech1983principles} Politeness Principle (PP). The Cooperative Principle is stated as \enquote{\textit{Make your conversational contribution such as is required, at the state at which it occurs, by the accepted purpose or direction of the talk exchange in which you are engaged}}. It comprises of four maxims of conversation, \textit{viz.} \enquote{maxim of quantity} (informative), \enquote{maxim of quality} (truthful), \enquote{maxim of relation} (relevant), and \enquote{maxim of manner} (clear), collectively known as the Gricean maxims. The Politeness Principle, on the other hand, is defined as \enquote{\textit{in its negative form, minimize (other things being equal) the expression of impolite beliefs and in the corresponding positive form, maximize (other things being equal) the expression of polite beliefs}}. It comprises of six maxims, \textit{viz.} \enquote{Tact maxim} (Minimize cost to other; Maximize benefit to other), \enquote{Generosity maxim} (Minimize benefit to self; Maximize cost to self), \enquote{Approbation maxim} (Minimize dispraise of other; Maximize praise of other), \enquote{Modesty maxim} (Minimize praise of self; Maximize dispraise of self), \enquote{Agreement maxim} (Minimize disagreement between self and other; Maximize agreement between self and other) and \enquote{Sympathy maxim} (Minimize antipathy between self and other; Maximize sympathy between self). These maxims are popularly known as politeness maxims. The PP trades off with the CP; indeed, it offers considerable explanatory power to the CP, in the sense that the CP explains \textit{how} individuals express indirect meanings, whereas the PP explains \textit{why} individuals express indirect meanings. To illustrate this, \citet{leech1992pragmatic} analyzed an excerpt from the G.B. Shaw’s play \enquote{\textit{You Never Can Tell}} which is as follows: \\
\begin{tcolorbox}[width=\linewidth, colback=white, colframe=black]
\scriptsize{[\textbf{Context:} The waiter, as the most tactful communicator, has been chosen to convey some
bad news to Crampton, namely that he is Philips’ father.]\\
\textbf{Waiter}: [smoothly melodious] Yes, sir. Great flow of spirits, sir. A vein of pleasantry, as you might say sir ... The young gentleman’s latest is that you’re his father.\\
\textbf{Crampton}: What! \\
\textbf{Waiter}: Only his joke, sir, his favourite joke. Yesterday I was to be his father ...}
\end{tcolorbox}
This excerpt demonstrates that the waiter is successful in communicating Crampton's parentage. Meanwhile, to retain politeness (i.e., to avoid upsetting), he further claims that this information is false and just a joke; in that way, he compromises the \enquote{maxim of quality}. 

Both CP and PP are exposed to criticism. The CP receives criticism owing to the fact that Gricean maxims do not apply universally, and PP is disputed by arguing that it has a large number of maxims which results in the overlap. Brown and Levinson, among others, have emphasized this aspect and formulated their theory of politeness, popularly known as Brown and Levinson's (B\&L) Politeness Theory \cite{brown1987politeness,brown1978universals}.

%\item 
\textbf{Face-based Approach to Politeness. }\citet{brown1978universals} (B\&L) Politeness Theory is the most well-known and researched in the field of politeness. B\&L studied politeness by utilizing \citet{goffman1967interaction} idea of face, which defines face as \enquote{the positive social value a person effectively claims for himself by the line others assume s/he has taken during a particular contact. Face is an image of self-delineated in terms of approved social attributes}. B\&L treated face as \textbf{basic wants}, which each member is aware that every other member wants, which are generally in each member's best interests to satisfy partially. They introduced two notions of face, namely \textit{positive face} and \textit{negative face}. Positive face reflects an individual's wants to be desirable to at least some others. Negative face, on the other hand, is the individual's desire that his actions be unimpeded by others. This theory is dependent on the hypothesis that certain kinds of speech acts threaten either the speaker's (S) face or the addressee's (H) face. Consequently, they define politeness as the necessary element to mitigate such face-threatening acts (FTAs). They claimed that different sociological variables, namely Relative Power (P, power dynamic between S and H, whether H is superior, subordinate, or at the same level as S), Distance (D, social distance between S and H, whether S and H are close friends, family members or colleague) and Absolute Ranking (R, degree of sensitivity of the topic in a particular situation or culture) determine the amount of face threat of a particular act. Eventually, \citet{brown1978universals} formulated five sets of super-strategies for doing FTAs that are associated with the degree of face threat, namely \textit{bald on-record, positive politeness, negative politeness, Off-record/indirect and withhold the FTA}. They mentioned in their politeness theory that bald on-record as a politeness strategy is a direct way of saying things to the hearer without making an effort to reduce the potential threat to the hearer's self-esteem. However, there are some methods to employ this strategy that implicitly mitigate FTAs, such as straightforwardly offering advice (e.g., \textit{Leave it, I will clean it up later.}). They stated that the frequent use of such a strategy may cause discomfort to the hearer; hence, it is typically employed in contexts where the speaker shares a strong bond with the hearer, such as within family or close friendships. They enumerated several scenarios in which one might opt for this strategy, including urgent situations (e.g. \textit{Watch out!}), situations when efficiency is necessary (e.g., \textit{Hear me out.}), and so on. Positive politeness strategies aim to mitigate the potential harm to the hearer's positive self-image. These strategies are employed to enhance the hearer's feeling of self-esteem, their interests, or their possessions. These are typically applied when there is a level of familiarity among the participants or when it is essential to address an individual's need for positive self-esteem. Negative politeness strategies, on the other hand, are directed towards the hearer's negative face and emphasize avoidance of imposition on the hearer. The off-record (indirect) strategy involves the use of indirect language and removes the speaker from the potential to be imposing. A few example sentences for each of these strategies are shown in Table \ref{tab:fta}. 

B\&L work primarily focuses on positive and negative politeness, which in turn, consists of 15 output strategies (i. Notice, Attend to H, ii. Exaggerate, iii. Intensify interest to H, iv. Use in-group identity markers, v. Seek agreement, vi. Avoid disagreement, vii. Presuppose/raise/assert common ground, viii. Joke, ix. Assert/presuppose S's knowledge of and concern for H's wants, x. Offer, promise, xi. Be optimistic, xii. Include both S and H in the activity, xiii. Give (or ask for) reasons, xiv. Assume or assert reciprocity, xv. Give gifts to H) and 10 output strategies (i. Be conventionally indirect, ii. Question, hedge, iii. Be pessimistic, iv. Minimize the imposition, v. Give deference, vi. Apologise, vii. Impersonalise S and H, viii. State the FTA as a general rule, ix. Nominalise, x. Go on-record as incurring a debt or as not indebting H), respectively \cite{brown1978universals}. Each output strategy serves as a method for fulfilling the strategic objectives of a higher-level super-strategy. %The different output strategies for both positive and negative politeness super-strategies as outlined by B\&L %, along with the examples 
\begin{table*}[hbt!]\scriptsize
  \caption{Strategies for doing FTAs. S refers to the speaker, and H refers to the addressee.}
  \label{tab:fta}
  \begin{adjustbox}{max width=\linewidth}
      \begin{tabular}{|l|l|l|}
\hline
\textbf{Strategy} &
  \textbf{Explanation} &
  \textbf{Example} \\ \hline
Bald on-record &
  \begin{tabular}[c]{@{}l@{}}Perform the FTA in the most direct, clear, unambiguous \\ and concise way possible \end{tabular} &
  \begin{tabular}[c]{@{}l@{}}(i) Please make a cup of tea. (said to a close friend) \\(ii) Don't forget to clean the blinds! \end{tabular}\\ \hline
Positive politeness &
  Minimizing the threat to H's positive face wants &
  \begin{tabular}[c]{@{}l@{}}(i) Yum! You make such a great sandwich! Would you make some?\\(ii) This piece of work is really fantastic. Well done! \end{tabular} \\ \hline
Negative politeness &
  Redressal of H's negative face wants &
  \begin{tabular}[c]{@{}l@{}}(i) Could you please serve the meal? \\(ii) I hope offense will not be taken.\end{tabular} \\ \hline
Off-record (indirect) &
  \begin{tabular}[c]{@{}l@{}}Perform the FTA such that there is more than one \\ unambiguously attributable intention so that the actor cannot \\ be held to have committed himself to one particular intent\end{tabular} &
  \begin{tabular}[c]{@{}l@{}}(i) If S says \enquote{I have stomachache.}, the H may infer that S is \\ asking for some medicines, however, if questioned, S may deny this.\\(ii) S: Are you going out? -H: Yes, but I’ll come home early.\end{tabular} \\ \hline
Withhold the FTA &
  The speaker makes a conscious decision to abstain from performing the FTA &
  - \\ \hline
\end{tabular}
\end{adjustbox}
\end{table*}

However, B\&L's theory of politeness is also exposed to criticism mainly because of its universality claim \cite{culpeper201113}. They argued that their conception of \enquote{face} applied across diverse cultures and languages. Moreover, they failed to articulate the role of context in the judgment of politeness; they assumed that certain speech acts/strategies are inherently polite and will produce the same politeness effects. While addressing these issues among others, several linguists posit discursive approaches to politeness.

%\item 
\textbf{Discursive Approach to Politeness. }The discursive approach to politeness is a noteworthy line of work in the field of linguistic politeness \cite{gino2001critique,locher2006power,locher2006polite,locher2005politeness,mills2003gender,watts2003politeness,watts20052}. \citet{locher2005politeness} made a distinction between first-order politeness (politeness as is perceived by the speakers of the language, the lay person's understandings) and second-order politeness (theoretical constructs of politeness model proposed in the literature) and claimed that there is no one meaning of the term `politeness', but rather a subject of discursive struggle. The discursive approach to politeness emphasizes the context and the situated and emergent meanings rather than pre-defined meanings. Furthermore, this approach acknowledges that politeness is evaluative in essence i.e., it is used to make judgments about people's behavior and is social norm-oriented, which provides insight into the concept of appropriateness. \citet{locher2006polite} stated that the \enquote{discursive approach to politeness stresses that we first of all have to establish the kind of relational work the interactants in question employ to arrive at an understanding of the then-current norms of interaction}. %The studies presented in  \cite{watts2003politeness,locher2006power,locher2005politeness} adopt the relational approach which is described in the next section.

%\item 
\textbf{Relational Approach to Politeness. }The relational approach to politeness is based on the concept of second-order politeness. \citet{christie2007relevance} pointed out that %that majority of the work states that 
politeness is some form of `relational work'. The term `relational work' relates to communication at the interpersonal level that helps in negotiating the relationships among people in interaction \cite{locher2005politeness,locher2008relational}. It encompasses the whole spectrum of behavior like rude, impolite, normal, polite, etc. As such, `relational work' cannot be simply limited to a distinction between polite and impolite behavior; in fact, there are some forms of `relational work', which is neither polite nor impolite. \citet{locher2005politeness} addressed facework from a new perspective, namely `relational work', by extending the research scope from \enquote{direct, impolite, rude, or aggressive interaction to polite interaction, encompassing both appropriate and inappropriate forms of social behavior} \cite{locher2006power}. Additionally, they proposed a lay interpretation of polite conduct as politic/appropriate verbal or non-verbal behavior in any social encounter and defined polite behavior as the surplus of politic behavior. This is illustrated in \citet{culpeper201113} using an example from \citet{watts2003politeness} (Example1) and a rework of this example (Example2). \\
% \textbf{Example1:} politic behaviour \\
% S: Would you like some more coffee? \\
% H: Yes, \textit{please} \\ 
% \textbf{Example2:} polite/positively marked behaviour \\
% S: Would you like some more coffee? \\
% H: Yes, \textit{please, that's very kind, coffee would be wonderful.} \\
\begin{minipage}{0.45\textwidth}
\begin{tcolorbox}[width=\linewidth,colback=white,colframe=black]
\textbf{Example1:} politic behaviour \\
S: Would you like some more coffee? \\
H: Yes, \textit{please}
\end{tcolorbox}
\end{minipage}
\hfill
\begin{minipage}{0.55\textwidth}
\begin{tcolorbox}[width=\linewidth,colback=white,colframe=black]
\textbf{Example2:} polite/positively marked behaviour \\
S: Would you like some more coffee? \\
H: Yes, \textit{please, that's very kind, coffee would be wonderful.}
\end{tcolorbox}
\end{minipage}\\
There are no separate distinctions between unmarked politic behavior and positively marked behavior, as they are of the same character and may overlap in specific contexts. This approach represents the premise that the understanding of polite and impolite behavior varies depending on contextual and social variables.

%\item 
\textbf{Frame-based Approach to Politeness. }
The authors in %\cite{terkourafi2001politeness,terkourafi2002politeness,terkourafi2005beyond,terkourafi2005pragmatic,terkourafigeneralised} 
\cite{terkourafi2001politeness,terkourafi2002politeness,terkourafi2005beyond,terkourafi2005pragmatic} introduced the frame-based approach to politeness. In these works, the authors asserted that the particular context of use and concrete linguistic realizations i.e., formulae, which together constitute \enquote{frames}, should be analyzed. This helps eliminate the problematic concept of directness. Furthermore, the authors in \cite{terkourafi2005beyond,terkourafi2005pragmatic,culpeper201113} stated that \enquote{[i]t is the regular co-occurrence of particular types of context and particular linguistic expressions as the unchallenged realizations of particular acts that create the perception of politeness}. The frame-based approach to politeness is data-driven, in contrast to the classic and post-modern views of politeness, which are theory-driven \cite{terkourafi2001politeness}. Data-driven means that it is predicated in the analysis of a large Cypriot Greek corpus, and it recognizes the norms to the level that these may be observed empirically. These norms were then evaluated quantitatively in order to identify regularities of co-occurrence between linguistic expressions and their extra-linguistic contexts of use. This approach provides a rich, coherent, and pragmatic account of politeness. It is mostly used by \cite{terkourafi2001politeness,terkourafi2002politeness,terkourafi2005beyond,terkourafi2005pragmatic} to investigate politeness in Cypriot Greek, and applying this approach to other contexts is another dimension of research in the field of linguistic politeness.  
%\end{enumerate}
% {The overview of the different facets of politeness considered in these approaches and their significant disagreements/criticisms in the linguistic studies are depicted in Figure \ref{pol_approaches}. These different approaches to politeness offer distinct perspectives on how people manage politeness in communication. They differ in their emphasis on linguistic maxims, face, discourse, relationships, and social frames. 

While researchers often disagree about which approach is most comprehensive or applicable in all contexts, each approach contributes valuable insights into understanding politeness in communication. The choice of approach may vary depending on the specific research question, context, and cultural considerations.

\subsection{Politeness Variations in Linguistic Studies}
Politeness analysis is a complex phenomenon because the perception of (im)politeness varies from one individual to another, and there are multiple gradations of more- or less- (im)polite behavior \cite{graham2007disagreeing}. Thus, politeness needs to be examined in a social context. Several factors influence the use of politeness in a particular situation. In this section, we will discuss such factors emphasizing the role of gender, age, and culture in politeness usage in interaction.

% \begin{enumerate}[listparindent=1.5em]
% \setlength\itemsep{0.5em}

%\item 
{\textbf{Politeness and Gender. }}There has been a good amount of research on language and gender traits that attempt to identify and explain the differences in speech styles of males and females. One of the most significant distinctions has been discovered in linguistic politeness. Many studies in socio-linguistics have observed the variations in the use of politeness strategies between males and females and concluded that linguistic politeness is influenced by gender \cite{lakoff1973language,brown1987politeness,mcelhinny1996language,mills2003gender}. According to these studies, women are more likely to use politeness strategies and compliments than men do while communicating. The work of \citet{lakoff1973language} is a groundbreaking study in addressing the problem of gender-related differences in politeness. The author claimed that women sound much more polite than men as they tend to use tag-questions more often than men. The tag-questions do not force belief or agreement on the addressee, and hence their use by women while interacting makes them sound more polite. Moreover, the author claimed that female speech includes many super-polite forms like \enquote{\textit{would you mind...}}, \enquote{\textit{would it be comfortable...}} etc., and consultative devices in utterances, for example, \enquote{\textit{could you tell what trouble you're facing?}}, \enquote{\textit{will you clean the window?}}. \citet{holmes2013women} agreed that, in general, females use more tags than males, as claimed by \citet{lakoff1973language}. Besides, the author also identified that females tend to consider question-tags to indicate politeness. Males, on the other other hand, use them to express uncertainty. \citet{holmes1988paying} stated that compliments are considered to be positive politeness strategies as they aim to strengthen the solidarity between the interlocutors. The notable gender difference in politeness is shown in how women and men utilize compliments. According to this study, women tend to give and receive more compliments than men. \citet{wardhaugh2006introduction} argued that females employ polite phrases like \enquote{\textit{please}}, \enquote{\textit{if you don’t mind}} etc. more often than males in conversations. \citet{montgomery1998multiple} claimed that out of age, social status, race, and gender, gender is the single most contributing factor to the occurrence of multiple modals. This study argued that both men and women are more sensitive to the face of women with whom they are conversing than to those of men. \citet{mills2003gender} provided a comprehensive examination of the assumptions underlying the relationship between gender and politeness. In this study, the author developed a dynamic conception of the association between gender and politeness and established a discourse-based analytical perspective on politeness. The author argued that politeness is a set of strategies rather than a choice of suitable utterances.

%\item 
\textbf{Politeness and Age. }Age is one of the significant aspects that might affect one’s views, experiences, social interactions, and linguistic engagements. Age-specific views of politeness hold that people of different age groups talk and behave differently as far as politeness is concerned. \citet{pfuderer1968scale} suggested that linguistic politeness is shown precisely when the listener is older or senior in the position. \citet{lakoff1973logic} claimed that adults must adhere to politeness norms and maxims. Researchers have investigated the influence of age on emotional well-being and emotional language. \citet{carstensen2011emotional} have shown that emotional well-being increases with age and that older people exhibit more positive emotions than younger ones. As politeness is an essential element of affective conversational behavior \cite{gupta2007rude}, recently, researchers have started investigating how age differences affect the choice of politeness strategies in interactions. Many researchers have focused on determining how and when children acquire politeness and how they use and comprehend the different polite forms while conversing. %According to \citet{ervin1977wait}, the knowledge of the linguistic form of polite requests and the knowledge of pragmatic requests rules within a given situational and social context are the two main factors affecting the generation and understanding of the polite register. These two abilities enable the children to make indirect requests such as \enquote{hints} \cite{mitchell1977pragmatics}. 
\citet{kuntay2014crosslinguistic} pointed out in their work that early studies on politeness focused chiefly on English-speaking children with an emphasis on the learning of politeness routines such as \enquote{\textit{thank you}}, \enquote{\textit{please}}, \enquote{\textit{I'm sorry}}, and comparable findings have been observed by the politeness studies with the children from other cultures like the one with Italian children \cite{bates1977social}, Japanese-speaking children \cite{nakamura2002polite}, to mention a few, demonstrating that they are also socialized into politeness norms at a younger age. Studies also suggest that the use of the word \enquote{\textit{please}} appears early, while indirect requests and politeness devices are used with increasing age. Moreover, the works by \citet{maratsos1973nonegocentric} and \citet{shatz1973development} showed that children are sensitive to their listeners as they use polite forms according to their needs and age. 

%\item 
\textbf{Politeness and Culture. } Politeness is a crucial aspect of communication, especially in cross-cultural contexts where miscommunications or misunderstandings may adversely affect the relationship between the interlocutors. The maxim-based approach to politeness introduced by \citet{leech2005politeness} accommodates a broader range of cultural variations. This seminal work is likely to be the most appropriate for East-West comparisons. \citet{lakoff1973logic} argued that politeness tends to avoid conflicts, and there are three rules for avoiding conflicts: camaraderie, deference, and distance. The authors mentioned that different cultures have different dominant rules; for example, Australian culture often emphasizes camaraderie, Japanese culture emphasizes deference, and British culture emphasizes distance. There are also linguistic differences in how politeness is expressed and perceived across cultures \cite{holtgraves1990politeness,janney1993universality}. For instance, there are significant differences in how politeness is conveyed and evaluated by the Chinese and Americans; what might be considered polite by the Chinese may sound utterly rude to a typical American. To illustrate, a cordial offer in Mandarin, \enquote{Come and eat with us, or I will get mad at you!} may sound extremely impolite to an American \cite{gu1990politeness}. Furthermore, there are differences between Chinese and Westerners in the ways of greeting and farewell, expressing thanks, praising, and in using addressing terms for others. For example, Westerners often use \enquote{Hi!} or \enquote{Hello!} to greet, while the Chinese prefer to ask \enquote{What are you doing here?} or \enquote{Have you eaten?}, which are perceived as interference with privacy by the Westerners and hence, considered impolite. Thus, understanding politeness across cultures is vital as it helps to enhance intercultural communication. 
%\end{enumerate}

\vspace{1em}
\noindent Some of these linguistic theories of politeness have a strong correlation with advancements in computational politeness. For instance, a seminal work by \citet{danescu2013computational} was an initial attempt to carry forward the groundwork done in discourse pragmatics, primarily by the \cite{lakoff1973logic,brown1987politeness}, in identifying and discussing social politeness. Moreover, the evaluative nature of politeness, necessary for making judgments about people’s behavior, is essentially the different forms of contextual information that computational politeness aims to capture. In general, politeness means socially correct and appropriate speech and behavior. However, what context is required to understand and evaluate politeness forms a crucial component. This contextual knowledge can be of various forms, like cultural, gender-specific, age-specific, and/or situational contexts. To illustrate, compare the polite example, `\textit{Hello! How are you?}' with another `\textit{Let's go for a cup of tea.}' The former is likely to be polite to all the people. The latter is likely to be polite for most people (friends, family, or long-standing colleagues). However, for the people who are new or senior in a position, let's say, one's employer, the statement is not polite. A direct statement like the latter one may threaten the hearer's face in restricting their freedom of action and thus may be perceived as impolite by them. Hence, the politeness understanding and computational approaches to politeness are contingent on what information or context is known.  

\subsection{Relationship with Etiquette, Formality, Emotion and Offensiveness}
Politeness is related to other forms of social language and behavior, such as etiquette, formality, and emotion. The terms \enquote{etiquette} and \enquote{politeness} are likely to conjure up images of formal curtsies and the proper dining fork to use. Although both the terms are often used interchangeably, there is a subtle difference between the two. Etiquette is the protocol via which we communicate politeness. It uses verbal, physical, gestural, and even more primitive forms of communication. For instance, people can show respect by posture, quiet speech, or by using titles and honorifics. Similar to politeness, there are differences in the use of etiquette codes across cultures \cite{miller2006computational}. %Cultural etiquette varies from one group to another, for example, between military and clerical employees or between marketplace negotiations in the Middle East and a shopping mall in the United States. 
Politeness and etiquette are very much at the forefront of managing social interactions and thus play a crucial role in training and predicting social interaction behaviors and perceptions across cultures.

Formality is another important dimension of stylistic variation in language \cite{biber1991variation,hudson199437}. %Consider the sentences, `\textit{Those suggestions were unsolicited and undesirable.}' and `\textit{That's the most ridiculous suggestion I've ever heard of.}' Even if both statements convey the same meaning, the first is somewhat more formal. 
Stylistic distinctions can frequently have a greater influence on how a listener interprets a statement than the literal meaning does \cite{hovy1987generating}, and the ability to recognize and respond to differences in formality is a necessary part of complete language understanding. \citet{heylighen1999formality} argued that formality %has close relations to informativeness and implicature. In particular, they stated that formality 
emerges out of a communicative objective - to maximize the amount of information being conveyed to the listener while at the same time maintaining (or at least appearing to maintain) Grice’s communicative maxims of Quality, Quantity, Relevance, and Manner as much as possible \cite{grice1975logic}. %They further suggested that the language should be more formal when addressing vast audiences or discussing abstract subjects, and the shared social context is less. Few authors use formality/informality to describe the characteristics of a social situation. The relevant characteristics of a situation may have something to do with a prevailing affective tone, so a formal situation requires a display of seriousness, politeness, and respect. For example, 
\citet{fischer1965stylistic} discussed the use of \enquote{respect vocabulary} and \enquote{formal etiquette} as the display of politeness marking a formal situation. Formality is closely linked with various stylistic dimensions, including politeness \cite{irvine1979formality,brown1979speech}, suggesting a positive correlation between them.
%Formality is often considered to be highly related with, and even to subsume, several other stylistic dimensions including but not limited to politeness \cite{irvine1979formality,brown1979speech}. These linguistic studies suggest that formality is usually positively correlated with politeness.

Another aspect of pragmatics that has gained importance in recent politeness studies %refers to 
is the emotional aspects \cite{langlotz2017politeness,renner2020directness}. \citet{brown1987politeness} mentioned displaying emotion or lack of control of emotions as positive politeness strategies or potentially face-threatening acts. The authors further referred to the `communication of affect' (display rules and emotional stance) as the display of affect, which is socially constructed, with cultural and situational expectations about what and how feelings should be displayed. They mentioned that communication of appropriate levels of affect is primarily concerned with linguistic politeness. \citet{culpeper2017palgrave}, in his description of requests, pointed out the significance of affect and familiarity as components of B\&L's \textit{distance} (D) variable. \citet{kienpointner2008impoliteness} argued that the variables, \textit{power} (P) and \textit{distance} (D), proposed by B\&L, also contain the elements of emotions. %The authors in \cite{chang2011strategic,kienpointner2008impoliteness,culpeper2010conventionalised,culpeper2011impoliteness,culpeper2014pragmatics,cashman2006impoliteness,locher2008relational} studied meta-comments about relational work that contain emotional cues and emotional assessments within interaction to gain a better understanding of breaches and maintenance of norms. 
\citet{culpeper2011impoliteness} asserted that \enquote{\textit{displaying emotions such as contempt or anger has nothing in itself to do with
impoliteness. However, somebody displaying great contempt for and anger at someone and doing so publicly may be judged (..) to have acted in an inappropriately and unfairly hurtful way (..), causing an emotional reaction such as embarrassment or anger.}} %In other words, it is the act of displaying strong emotions, and the emotional consequences that are elicited on the side of the interlocutor(s) are significant factors for assessing (im)politeness. 
%Likewise, \citet{langlotz2017politeness} highlighted that the emotional reactions of the interlocutors are primarily concerned with theorizing how emotions cognitively contribute to relational understandings. %Politeness has also been shown to lead to the development of rapport \cite{spencer2005politeness}. In the discussion on rapport management, \citet{spencer2005politeness} illustrated that certain emotional reactions such as \textit{joy}, \textit{anger}, \textit{sadness}, \textit{surprise} and their sub-groups of S and H play an important role in negotiation of face concerns. %\citet{blitvich2013introduction} further argued that concepts of both face and identity should be linked to emotions. 
Numerous linguistic studies indicate that emotions are essential in appraisals and relational work. Therefore, investigating the relationship between emotions and politeness can benefit linguistic politeness research.%that emotions are a fundamental part of appraisals and of relating and assessing relational work. Thus, exploring the interconnectedness between emotions and politeness can be advantageous for linguistic politeness research.

Politeness is often discussed in connotation with another linguistic aspect, offensiveness. Politeness and offensiveness are essential aspects of linguistic communication that influence interpersonal relationships and social dynamics. Politeness encompasses various linguistic strategies employed to convey respect, courtesy, and consideration for others, facilitating a positive and harmonious atmosphere. Offensiveness, on the other hand, pertains to the use of language that is disrespectful, rude, or hurtful to others. The notion of offensiveness lies at the core of impoliteness \cite{culpeper2011impoliteness}, and it can take various forms from general and often harmless profanity to derogatory and abusive language intended to cause harm and emotional distress, such as cyberbullying and hate speech \cite{waseem2017understanding}. \citet{bousfield2008impoliteness} and \citet{culpeper2011impoliteness} delved into the concept of impoliteness, equating the intent to offend with posing a threat or causing damage to one's face. \citet{culpeper2011impoliteness,culpeper2016impoliteness}, and \citet{haugh2010email} proposed a theoretical framework to explain impoliteness, considering it as an attitudinal stance of the speaker. Offense, on the other hand, can be perceived as emotional reactions experienced by recipients or the cause of such reactions. It's important to note that acting impolitely doesn't always equate to offending, as the perception of offense varies among interlocutors. For example, interrupting a teammate during a presentation with disruptive comments may be seen as impolite, but using derogatory language to insult their intelligence is considered offensive. Offensiveness, therefore, can be viewed as an extreme form of impoliteness involving a direct attack on someone's dignity or identity. % \citet{bousfield2008impoliteness} regarded the speaker's intent to `\textit{offend}' equivalent to posing a threat or causing damage to one's face. \citet{culpeper2011impoliteness} and \citet{culpeper2016impoliteness} formulated a theoretical framework to explain the concepts of impoliteness and the act of causing offense. He construed impoliteness as a particular attitudinal stance on the part of speakers. In contrast, the offense was comprehensively examined as either (a) an emotional reaction experienced by recipients that can range in intensity (such as feelings of anger, dissatisfaction, or irritation triggered by an offending incident), or (b) the origin of such emotional responses (i.e., a cause of feelings of anger, dissatisfaction, or irritation). \citet{culpeper2011impoliteness}, \citet{culpeper2016impoliteness}, and \citet{haugh2010email} argued that acting impolitely does not necessarily equate to offending and that the interlocutors may or may not perceive offense when exposed to behavior or speech that appears impolite. For instance, during a presentation, if a person A interrupts his/her teammate B multiple times by making comments like \enquote{\textit{This idea is not going to work}}, or \enquote{\textit{You are missing the point}}, etc., then A's behavior will be perceived as impolite because A is being disruptive and disrespectful in the way he is communicating his disagreement. However, if A interrupts by passing personal comments like \enquote{\textit{Your ideas both are shit. I don't believe how someone can come up with such nonsense.}}, then it is something beyond impoliteness. It is considered offensive because it attacks B personally, using derogatory language to insult B's intelligence. In essence, offensiveness can be perceived as a more extreme form of impoliteness, as it usually involves a direct attack on someone's dignity or identity. 

Nevertheless, it cannot be denied that instances of impoliteness/offensiveness are closely linked to the concept of politeness \cite{culpeper201113}. The degree of impoliteness or offensiveness often varies depending on the expected level of politeness in a given situation. For example, asking the Vice Chancellor of a university to be quiet might be more offensive than making the same request to a young daughter. %Sarcasm also illustrates this relationship, as it involves a trade-off with politeness. For instance, saying "thank you" sarcastically after experiencing disappointment highlights the disparity between the expected polite response and the actual sentiment. 
\citet{vogel2015some} mentioned that behaviors of politeness and impoliteness can modulate offensiveness. They exist on a continuum of communication behaviors and can coexist within the same interaction. A message may have a polite tone but still convey an offensive message or criticism. Conversely, impolite language may not necessarily intend to be offensive; it could stem from a lack of awareness of social norms or an expression of frustration rather than an intentional attempt to harm. The relationship between politeness and offensiveness is multifaceted, influenced by context, intent, and individual perception \cite{culpeper201113}; what one person deems polite, another may consider impolite or offensive, highlighting the subjective nature of these concepts.

These characteristics of politeness pave the way for a few works in computational politeness research, which we will discuss in forthcoming sections.

\section{Problem Definition}
\label{section3}

%%%%%%%%%%%%%%%%%%%%%%%%%%%%%%%%%%%%%%%%%%%%%%%%%%%%
% The increasing completeness and approaching perfection of natural language processing tasks that examine language at a more granular level along with the advancement in machine learning and deep learning techniques has resulted in an increased emphasis on computational discourse pragmatics. Politeness is one of the widely studied topics in discourse pragmatics. Computational models and simulations of politeness-oriented social interactions are useful for a variety of downstream applications such as customer behavior modeling, response generation for conversational agents and similar intelligent systems like robots, understanding behavior of online communities, and language learning, to mention a few. Computational approaches to politeness are mainly researched under two broad headings: identification and generation, which we will discuss in subsequent sub-sections.
The advancements in machine learning and deep learning techniques profoundly impact the performance of various Natural Language Processing (NLP) tasks, such as Part-of-Speech tagging, Named Entity Recognition, and Semantic Parsing, to name a few. These techniques have also demonstrated remarkable success in contextual language understanding and generation, enabling breakthroughs in computational discourse pragmatics. Politeness is one of the widely studied topics in discourse pragmatics. Computational models and simulations of politeness-oriented interactions hold significant utility across various applications. These include modeling customer behavior, generating responses for conversational agents and similar intelligent systems like robots, comprehending the dynamics of online communities, and facilitating language learning, among others. In computational politeness, research predominantly falls into two overarching categories: identification and generation, which we will discuss in subsequent sub-sections. Table \ref{problemdef_detect} and Table \ref{problemdef_generate} provide a summary of past works in computational politeness identification and politeness generation, respectively. While several interesting conclusions may be drawn from both the tables, the three notable ones are: \textbf{(i).} requests are the pre-dominant text form for computational politeness analysis; \textbf{(ii).} politeness is imperative in both conversational (open-domain and task-oriented) and non-conversational settings for promoting user experience, and \textbf{(iii).}
incorporation of contextual information in the form of conversational history is a recent trend in computational politeness modeling for dialogue systems.

\begin{table}[hbt!]\scriptsize
    \centering
    \caption{Summary of computational approaches to politeness identification along different parameters.}
    \begin{adjustbox}{max width=\linewidth}
\renewcommand{\arraystretch}{1.9}
    \begin{tabular}{|l|cc|cc|ccc|cc|cccc|cc|}

\hline
 & \multicolumn{2}{c|}{\textbf{Datasets}}                                                           & \multicolumn{2}{c|}{\textbf{Languages}} & \multicolumn{3}{c|}{\textbf{Approach}}                                                                                                                                                                                                                 & \multicolumn{2}{c|}{\textbf{Annotation}}                                       & \multicolumn{4}{c|}{\textbf{Features}}                                                                                                                        & \multicolumn{2}{c|}{\textbf{Context}}                                             \\ \hline %\cline{1-3} \cline{5-15} 
 & \multicolumn{1}{c|}{\textbf{\begin{turn}{90}Non-conversational\end{turn}}} &  \multicolumn{1}{c|}{\textbf{\begin{turn}{90}Conversational\end{turn}}} & %\multicolumn{1}{c|}{}
 
 \multicolumn{1}{c|}{\textbf{\begin{turn}{90}English\end{turn}}} & \multicolumn{1}{c|}{\textbf{\begin{turn}{90}Other\end{turn}}} &
 
 \multicolumn{1}{c|}{\textbf{\begin{turn}{90}Supervised\end{turn}}} & \multicolumn{1}{c|}{\textbf{\begin{turn}{90}\begin{tabular}[c]{@{}c@{}}(Semi/Weakly/Un)-\\Supervised\end{tabular}\end{turn}}} & \multicolumn{1}{c|}{\textbf{\begin{turn}{90}\begin{tabular}[c]{@{}c@{}}Reinforcement\\ Learning\end{tabular}\end{turn}}} & \multicolumn{1}{c|}{\textbf{\begin{turn}{90}Manual\end{turn}}} & \multicolumn{1}{c|}{\textbf{\begin{turn}{90}Automatic\end{turn}}} & \multicolumn{1}{c|}{\textbf{\begin{turn}{90}N-grams\end{turn}}} & \multicolumn{1}{c|}{\textbf{\begin{turn}{90}Lexical\end{turn}}} & \multicolumn{1}{c|}{\textbf{\begin{turn}{90}Syntactic\end{turn}}} & \multicolumn{1}{c|}{\textbf{\begin{turn}{90}Other\end{turn}}} & \multicolumn{1}{c|}{\textbf{\begin{turn}{90}Conversation\end{turn}}} & \multicolumn{1}{c|}{\textbf{\begin{turn}{90}Other\end{turn}}} \\ \hline
 
\citet{alexandrov2008regression} & \multicolumn{1}{c|}{}   &                     
\ding{51} & 
\multicolumn{1}{c|}{} & 
\ding{51} &   
\multicolumn{1}{c|}{{\ding{51}}}  & 
\multicolumn{1}{c|}{}  &                      & 
\multicolumn{1}{c|}{}               & 
                  & 
\multicolumn{1}{c|}{}                & \multicolumn{1}{c|}{\ding{51}}                & \multicolumn{1}{c|}{\ding{51}}                  & 
             & 
\multicolumn{1}{c|}{}                     & 
              
\\ \hline

 \citet{danescu2013computational} & \multicolumn{1}{c|}{\ding{51}}   &                     
  & 
\multicolumn{1}{c|}{\ding{51}} & 
  &   
\multicolumn{1}{c|}{{\ding{51}}}  & 
\multicolumn{1}{c|}{}  &                                  & 
\multicolumn{1}{c|}{\ding{51}}               & 
                 & 
\multicolumn{1}{c|}{\ding{51}}                & \multicolumn{1}{c|}{\ding{51}}                & \multicolumn{1}{c|}{\ding{51}}                  & 
              & 
\multicolumn{1}{c|}{}                     & 
               
\\ \hline
\citet{li2020studying} & \multicolumn{1}{c|}{\ding{51}}   &                     
& 
\multicolumn{1}{c|}{\ding{51}} & 
\ding{51}  &   
\multicolumn{1}{c|}{{\ding{51}}}  & 
\multicolumn{1}{c|}{}  &                                 & 
\multicolumn{1}{c|}{\ding{51}}               & 
                & 
\multicolumn{1}{c|}{}                & \multicolumn{1}{c|}{\ding{51}}                & \multicolumn{1}{c|}{\ding{51}}                  & 
             & 
\multicolumn{1}{c|}{}                     & 
             
\\ \hline
\citet{aubakirova2016interpreting} & \multicolumn{1}{c|}{\ding{51}}   &                     
  & 
\multicolumn{1}{c|}{\ding{51}} & 
  &   
\multicolumn{1}{c|}{{\ding{51}}}  & 
\multicolumn{1}{c|}{}  &                                  & 
\multicolumn{1}{c|}{}               & 
                  & 
\multicolumn{1}{c|}{}                & \multicolumn{1}{c|}{}                & \multicolumn{1}{c|}{}                  & 
              & 
\multicolumn{1}{c|}{}                     & 
              
\\ \hline
\citet{chhaya2018frustrated} & \multicolumn{1}{c|}{\ding{51}}   &                     
 & 
\multicolumn{1}{c|}{\ding{51}} & 
  &   
\multicolumn{1}{c|}{{\ding{51}}}  & 
\multicolumn{1}{c|}{}  &                                  & 
\multicolumn{1}{c|}{\ding{51}}               & 
                 & 
\multicolumn{1}{c|}{}                & \multicolumn{1}{c|}{\ding{51}}                & \multicolumn{1}{c|}{\ding{51}}                  & 
             & 
\multicolumn{1}{c|}{}                     & 
              
\\ \hline
\citet{kumar2021towards} & \multicolumn{1}{c|}{\ding{51}}   &                     
  & 
\multicolumn{1}{c|}{} & 
\ding{51}  &   
\multicolumn{1}{c|}{{\ding{51}}}  & 
\multicolumn{1}{c|}{}  &                                 & 
\multicolumn{1}{c|}{\ding{51}}               & 
                 & 
\multicolumn{1}{c|}{\ding{51}}                & \multicolumn{1}{c|}{}                & \multicolumn{1}{c|}{}                  & 
\ding{51}  & 
\multicolumn{1}{c|}{}    & 
              
\\ \hline

\citet{mishra2022predicting} & \multicolumn{1}{c|}{}   &                     
\ding{51} & 
\multicolumn{1}{c|}{\ding{51}} & 
&   
\multicolumn{1}{c|}{{\ding{51}}}  & 
\multicolumn{1}{c|}{}  &                           & 
\multicolumn{1}{c|}{\ding{51}}               & 
                 \ding{51} & 
\multicolumn{1}{c|}{}                & \multicolumn{1}{c|}{}                & \multicolumn{1}{c|}{}                  & 
             & 
\multicolumn{1}{c|}{\ding{51}}                     & 
              
\\ \hline

\citet{dasgupta2023graph} & \multicolumn{1}{c|}{\ding{51}}   &                     
 & 
\multicolumn{1}{c|}{\ding{51}} & 
&   
\multicolumn{1}{c|}{{\ding{51}}}  & 
\multicolumn{1}{c|}{}  &                           & 
\multicolumn{1}{c|}{}               & 
                  & 
\multicolumn{1}{c|}{}                & \multicolumn{1}{c|}{}                & \multicolumn{1}{c|}{}                  & 
             & 
\multicolumn{1}{c|}{}                     & 
              
\\ \hline

\citet{priya2023multi} & \multicolumn{1}{c|}{}   &                     
 \ding{51} & 
\multicolumn{1}{c|}{\ding{51}} & 
&   
\multicolumn{1}{c|}{{\ding{51}}}  & 
\multicolumn{1}{c|}{}  &                           & 
\multicolumn{1}{c|}{\ding{51}}               & 
                  & 
\multicolumn{1}{c|}{}                & \multicolumn{1}{c|}{}                & \multicolumn{1}{c|}{}                  & 
             & 
\multicolumn{1}{c|}{\ding{51}}                     & 
              
\\ \hline

\end{tabular}
\end{adjustbox}  
    \label{problemdef_detect}
\end{table}

%%%%%%%%%%%%%%%%%%%%%%%%%%%%%%%%%%%%%%%%%%%%%%%%%%%%

\subsection{Politeness as Natural Language Understanding Task}
Recently, a few research works on computational politeness have been inclined toward identifying various linguistic features of politeness and predicting whether the given text is polite or not. Thus, given a sentence, \enquote{\textit{Would you please help me?}} should be predicted as polite, while the sentence \enquote{\textit{Would you please stop?}} should be predicted as impolite. It is important to note that though both sentences contain the phrase `Would you please', which is generally considered a polite way to request or ask for something, both are not polite. The perception of politeness or impoliteness in these sentences depends on the context and the nature of the request. The former sentence is typically perceived as polite because it is a polite request for assistance. In this sentence, `Would you please', signifies respect and deference towards the person from whom the help is needed. In the latter sentence, though the polite marker `Would you please' is used, it is perceived as impolite. It is a request for someone to cease or desist from a particular action. If used with a harsh tone, it exacerbates the perception of impoliteness. Table \ref{problemdef_detect} provides a brief overview of the different datasets, languages, approaches, challenges, etc., on natural language understanding of politeness. %The past work presents a variety of possibilities for politeness labels in politeness identification task. For instance, 

The works by \citet{danescu2013computational} and \citet{aubakirova2016interpreting} consider two labels, \textit{viz.} polite and impolite for the classifier. Lately, \citet{mishra2022please} and \citet{mishra2022predicting} consider fine-grained labels for the classifier as: polite, somewhat\_polite, somewhat\_impolite, and impolite to effectively model the politeness variations in goal-oriented interactions. Thus, instead of simply classifying the given text as polite or impolite, they perform the fine-grained classification as \textit{\enquote{What is your departure city?}} - impolite, \textit{\enquote{Can you provide your departure city?}} - somewhat\_impolite, \textit{\enquote{Please, provide your departure city?}} - somewhat\_polite, \textit{\enquote{For further processing, I would like to know your departure city, please, name it?}} - polite. 

\citet{chhaya2018frustrated} formulate the task of politeness prediction as a regression problem and a 2-class classification problem in which the model predicts whether the text has politeness or not. \citet{yeomans2018politeness} develop a package in R that provides functions to extract politeness markers in the English language, graphically compare these markers to covariates of interest, develop a supervised model to identify politeness in
new documents and inspect high- and low-politeness documents. \citet{aljanaideh2020contextualized} model the problem of politeness detection in natural language requests as a clustering task. The authors have created a set of clusters for every word in the request. Each cluster contains contextualized representations of the word (obtained using a pre-trained BERT model) based on the context in which the word appears and the labels of items the word occurs in. The authors state that examining the context of using words helps enrich the linguistic analyses of the politeness task. For example, the hedge \enquote{\textit{Maybe}} can be used as a part of a polite request (\enquote{\textit{Maybe we can meet}}) or an impolite one (\enquote{\textit{Maybe it would be better if we never talked directly to each other ever again}}). \citet{voigt2017language} approach the task of politeness detection analogous to detecting respectfulness that police officers display to community members. The authors endeavor to predict various facets of respectful language (apologizing, giving agency, softening of commands, etc.) by analyzing linguistic features derived from theories of politeness, power, and social distance and find that particular forms of politeness are implicated in perceptions of respect. 

Lately, a few works delve into the dynamics of online interactions, exploring various factors beyond but associated with the politeness that contributes to positive conversational outcomes, such as constructive comments \cite{napoles2017automatically,napoles2017finding}, supportiveness \cite{buechel2018modeling}, receptiveness \cite{minson2020won}, or empathy \cite{buechel2018modeling,sharma2020computational,zhou2020condolence}. The authors in \cite{zhang2018conversations,bao2021conversations} define a novel task of forecasting the future behavior of a conversation from early signals. \citet{bao2021conversations} investigate the effect of prosocial behavior, including activities like helping, sharing, comforting, rescuing, and cooperating in online discussions. Inspired by the prosocial behavior theories in social psychology \cite{brown1991self}, the authors delineate eight broad categories of behavior, \textit{viz.} \textit{information sharing}, \textit{gratitude}, \textit{esteem enhancement}, \textit{social support}, \textit{social cohesion}, \textit{fundraising and donating}, \textit{mentoring}, and \textit{absence of antisocial behavior}. They attempt to forecast the prosocial quality of a conversation, demonstrating that these outcomes can be accurately forecasted from cues early in the conversation. \citet{zhang2018conversations} introduce the conversation prompts used to initiate different types of conversations in online collaborative settings. These prompts include \textit{factual check}, \textit{moderation}, \textit{coordination}, \textit{casual remark}, \textit{action statement}, and \textit{opinion}. The authors explore the correlation between these conversation prompts and predefined politeness strategies to anticipate potential conversation failures. A handful of studies investigate how politeness influences interactions involving individuals with opposing or conflicting views. \citet{jeong2019communicating} endeavor to discern the impact of politeness and respect \cite{danescu2013computational} on warm and friendly communication styles, exploring how these factors influence perceptions and interactions in distributive negotiation. \citet{yeomans2020conversational} aim to ascertain whether employing a more receptive communication style facilitates the pursuit of cooperative objectives among individuals with differing opinions, such as a readiness to collaborate in the future, interpersonal trust, and conflict de-escalation. They address conversational receptiveness by identifying the politeness cues that individuals perceive when assessing a counterpart's openness to dialogue. They utilize these cues to create an intervention to encourage conversation partners to communicate their willingness to engage with opposing views.

%%%%%%%%%%%%%%%%%%%%%%%%%%%%%%%%%%%%%%%%%%%%%%%%%%%%%%%%%%%%%%%%%%%%%%%%%%%%%%%%%%%%%%%%%%%%%%%%%%%%
\begin{table}[hbt!]\scriptsize
    \centering
    \caption{Summary of computational approaches to politeness generation along different parameters.}
    \begin{adjustbox}{max width=\linewidth}
\renewcommand{\arraystretch}{1.9}
    \begin{tabular}{|c|lc|lc|llc|lc|lllc|lc|}

\hline
 & \multicolumn{2}{c|}{\textbf{Datasets}}                                                           & \multicolumn{2}{c|}{\textbf{Languages}} & \multicolumn{3}{c|}{\textbf{Approach}}                                                                                                                                                                                                                 & \multicolumn{2}{c|}{\textbf{Annotation}}                                       & \multicolumn{4}{c|}{\textbf{Features}}                                                                                                                        & \multicolumn{2}{c|}{\textbf{Context}}                                             \\ \hline %\cline{1-3} \cline{5-15} 
 & \multicolumn{1}{c|}{\textbf{\begin{turn}{90}Non-conversational\end{turn}}} &  \multicolumn{1}{c|}{\textbf{\begin{turn}{90}Conversational\end{turn}}} & %\multicolumn{1}{c|}{}
 
 \multicolumn{1}{c|}{\textbf{\begin{turn}{90}English\end{turn}}} & \multicolumn{1}{c|}{\textbf{\begin{turn}{90}Other\end{turn}}} &
 
 \multicolumn{1}{c|}{\textbf{\begin{turn}{90}Supervised\end{turn}}} & \multicolumn{1}{c|}{\textbf{\begin{turn}{90}\begin{tabular}[c]{@{}c@{}}(Semi/Weakly/Un)-\\Supervised\end{tabular}\end{turn}}} & \multicolumn{1}{c|}{\textbf{\begin{turn}{90}\begin{tabular}[c]{@{}c@{}}Reinforcement\\ Learning\end{tabular}\end{turn}}} & \multicolumn{1}{c|}{\textbf{\begin{turn}{90}Manual\end{turn}}} & \multicolumn{1}{c|}{\textbf{\begin{turn}{90}Automatic\end{turn}}} & \multicolumn{1}{c|}{\textbf{\begin{turn}{90}N-grams\end{turn}}} & \multicolumn{1}{c|}{\textbf{\begin{turn}{90}Lexical\end{turn}}} & \multicolumn{1}{c|}{\textbf{\begin{turn}{90}Syntactic\end{turn}}} & \multicolumn{1}{c|}{\textbf{\begin{turn}{90}Other\end{turn}}} & \multicolumn{1}{c|}{\textbf{\begin{turn}{90}Conversation\end{turn}}} & \multicolumn{1}{c|}{\textbf{\begin{turn}{90}Other\end{turn}}} \\ \hline
 
\citet{madaan2020politeness} & \multicolumn{1}{c|}{\ding{51}}   &                     
  & 
\multicolumn{1}{c|}{\ding{51}} & 
  &   
\multicolumn{1}{c|}{{\ding{51}}}  & 
\multicolumn{1}{c|}{}  &                                  & 
\multicolumn{1}{c|}{}               & 
\ding{51}                 & 
\multicolumn{1}{c|}{\ding{51}}                & \multicolumn{1}{c|}{}                & \multicolumn{1}{c|}{}                  & 
            & 
\multicolumn{1}{c|}{}                     & 
               
\\ \hline
\citet{fu2020facilitating} & \multicolumn{1}{c|}{\ding{51}}   &                     
  & 
\multicolumn{1}{c|}{\ding{51}} & 
 &   
\multicolumn{1}{c|}{{}}  & 
\multicolumn{1}{c|}{\ding{51}}  &                                  & 
\multicolumn{1}{c|}{}               & 
                  & 
\multicolumn{1}{c|}{}                & \multicolumn{1}{c|}{}                & \multicolumn{1}{c|}{}                  & 
             & 
\multicolumn{1}{c|}{}                     & 
              
\\ \hline
\citet{niu2018polite} & \multicolumn{1}{c|}{\ding{51}}   &                     
 \ding{51} & 
\multicolumn{1}{c|}{\ding{51}} & 
&   
\multicolumn{1}{c|}{{}}  & 
\multicolumn{1}{c|}{\ding{51}}  &                         \ding{51}  & 
\multicolumn{1}{c|}{}               & 
                  & 
\multicolumn{1}{c|}{}                & \multicolumn{1}{c|}{}                & \multicolumn{1}{c|}{}                  & 
             & 
\multicolumn{1}{c|}{}                     & 
              
\\ \hline

\citet{sennrich2016controlling} & \multicolumn{1}{c|}{\ding{51}}   &                     
& 
\multicolumn{1}{c|}{\ding{51}} & 
\ding{51} &   
\multicolumn{1}{c|}{{\ding{51}}}  & 
\multicolumn{1}{c|}{}  &                       & 
\multicolumn{1}{c|}{}               & 
               \ding{51}   & 
\multicolumn{1}{c|}{}                & \multicolumn{1}{c|}{}                & \multicolumn{1}{c|}{}                  & 
             & 
\multicolumn{1}{c|}{}                     & 
              
\\ \hline

\citet{feely2019controlling} & \multicolumn{1}{c|}{\ding{51}}   &                     
  & 
\multicolumn{1}{c|}{\ding{51}} & 
\ding{51} &   
\multicolumn{1}{c|}{{\ding{51}}}  & 
\multicolumn{1}{c|}{}  &                           & 
\multicolumn{1}{c|}{}               & 
                 \ding{51} & 
\multicolumn{1}{c|}{}                & \multicolumn{1}{c|}{}                & \multicolumn{1}{c|}{}                  & 
             & 
\multicolumn{1}{c|}{}                     & 
              
\\ \hline
\citet{golchha2019courteously} & \multicolumn{1}{c|}{}   &                     
 \ding{51} & 
\multicolumn{1}{c|}{\ding{51}} & 
&   
\multicolumn{1}{c|}{{}}  & 
\multicolumn{1}{c|}{}  &   \ding{51}                        & 
\multicolumn{1}{c|}{\ding{51}}               & 
                  & 
\multicolumn{1}{c|}{}                & \multicolumn{1}{c|}{}                & \multicolumn{1}{c|}{}                  & 
             & 
\multicolumn{1}{c|}{\ding{51}}                     & \ding{51}
              
\\ \hline

\citet{firdaus2020incorporating} & \multicolumn{1}{c|}{}   &                     
 \ding{51} & 
\multicolumn{1}{c|}{\ding{51}} & 
\ding{51} &   
\multicolumn{1}{c|}{{}}  & 
\multicolumn{1}{c|}{}  &   \ding{51}                        & 
\multicolumn{1}{c|}{\ding{51}}               & 
                  & 
\multicolumn{1}{c|}{}                & \multicolumn{1}{c|}{}                & \multicolumn{1}{c|}{}                  & 
             & 
\multicolumn{1}{c|}{\ding{51}}                     & \ding{51}
              
\\ \hline

\citet{wang2020can} & \multicolumn{1}{c|}{\ding{51}}    &                     
  & 
\multicolumn{1}{c|}{\ding{51}} & 
&   
\multicolumn{1}{c|}{{\ding{51}}}  & 
\multicolumn{1}{c|}{}  &                          & 
\multicolumn{1}{c|}{}               & 
                 \ding{51} & 
\multicolumn{1}{c|}{}                & \multicolumn{1}{c|}{}                & \multicolumn{1}{c|}{}                  & 
             & 
\multicolumn{1}{c|}{}                     & 
              
\\ \hline

\citet{firdaus2022being} & \multicolumn{1}{c|}{}   &                     
 \ding{51} & 
\multicolumn{1}{c|}{\ding{51}} & 
&   
\multicolumn{1}{c|}{{\ding{51}}}  & 
\multicolumn{1}{c|}{}  &      \ding{51}                     & 
\multicolumn{1}{c|}{\ding{51}}               & 
                  & 
\multicolumn{1}{c|}{}                & \multicolumn{1}{c|}{}                & \multicolumn{1}{c|}{}                  & 
             & 
\multicolumn{1}{c|}{\ding{51}}                     & 
              
\\ \hline

\citet{mishra2022please} & \multicolumn{1}{c|}{}   &                     
 \ding{51} & 
\multicolumn{1}{c|}{\ding{51}} & 
&   
\multicolumn{1}{c|}{{}}  & 
\multicolumn{1}{c|}{}  &   \ding{51}                        & 
\multicolumn{1}{c|}{\ding{51}}               & 
                 \ding{51} & 
\multicolumn{1}{c|}{}                & \multicolumn{1}{c|}{}                & \multicolumn{1}{c|}{}                  & 
             & 
\multicolumn{1}{c|}{\ding{51}}                     & 
              
\\ \hline

\citet{silva2022polite} & \multicolumn{1}{c|}{}   &                     
 \ding{51} & 
\multicolumn{1}{c|}{\ding{51}} & 
&   
\multicolumn{1}{c|}{{\ding{51}}}  & 
\multicolumn{1}{c|}{}  &   \ding{51}                        & 
\multicolumn{1}{c|}{}               & 
                  & 
\multicolumn{1}{c|}{}                & \multicolumn{1}{c|}{}                & \multicolumn{1}{c|}{}                  & 
             & 
\multicolumn{1}{c|}{}                     & 
              
\\ \hline
%It's approach is unsupervised
\citet{saha2022countergedi} & \multicolumn{1}{c|}{\ding{51}}    &                     
  & 
\multicolumn{1}{c|}{\ding{51}} & 
&   
\multicolumn{1}{c|}{{}}  & 
\multicolumn{1}{c|}{\ding{51}}  &                          & 
\multicolumn{1}{c|}{}               & 
                  & 
\multicolumn{1}{c|}{}                & \multicolumn{1}{c|}{}                & \multicolumn{1}{c|}{}                  & 
             & 
\multicolumn{1}{c|}{}                     & 
              
\\ \hline

\citet{firdaus2022polise} & \multicolumn{1}{c|}{}    &  \ding{51}                   
  & 
\multicolumn{1}{c|}{\ding{51}} & 
&   
\multicolumn{1}{c|}{{}}  & 
\multicolumn{1}{c|}{}  &   \ding{51}                       & 
\multicolumn{1}{c|}{\ding{51}}               & 
                  & 
\multicolumn{1}{c|}{}                & \multicolumn{1}{c|}{}                & \multicolumn{1}{c|}{}                  & 
             & 
\multicolumn{1}{c|}{\ding{51}}                     & 
              
\\ \hline

\citet{mishra2022pepds} & \multicolumn{1}{c|}{}    &  \ding{51}                   
  & 
\multicolumn{1}{c|}{\ding{51}} & 
&   
\multicolumn{1}{c|}{{}}  & 
\multicolumn{1}{c|}{}  &   \ding{51}                       & 
\multicolumn{1}{c|}{}               & \ding{51}
                  & 
\multicolumn{1}{c|}{}                & \multicolumn{1}{c|}{}                & \multicolumn{1}{c|}{}                  & 
             & 
\multicolumn{1}{c|}{\ding{51}}                     & 
              
\\ \hline

\citet{mukherjee2023polite} & \multicolumn{1}{c|}{}    &  \ding{51}                   
  & 
\multicolumn{1}{c|}{\ding{51}} & 
&   
\multicolumn{1}{c|}{{\ding{51}}}  & 
\multicolumn{1}{c|}{}  &                         & 
\multicolumn{1}{c|}{}               & 
                  & 
\multicolumn{1}{c|}{}                & \multicolumn{1}{c|}{}                & \multicolumn{1}{c|}{}                  & 
             & 
\multicolumn{1}{c|}{\ding{51}}                     & 
              
\\ \hline

\citet{mishra2023pal} & \multicolumn{1}{c|}{}    &  \ding{51}                   
  & 
\multicolumn{1}{c|}{\ding{51}} & 
&   
\multicolumn{1}{c|}{{}}  & 
\multicolumn{1}{c|}{}  &  \ding{51}                       & 
\multicolumn{1}{c|}{}               & \ding{51}
                  & 
\multicolumn{1}{c|}{}                & \multicolumn{1}{c|}{}                & \multicolumn{1}{c|}{}                  & 
             & 
\multicolumn{1}{c|}{\ding{51}}                     & 
              
\\ \hline

\citet{ijcai2023p686} & \multicolumn{1}{c|}{}    &  \ding{51}                   
  & 
\multicolumn{1}{c|}{\ding{51}} & 
&   
\multicolumn{1}{c|}{{}}  & 
\multicolumn{1}{c|}{}  &  \ding{51}                       & 
\multicolumn{1}{c|}{}               & \ding{51}
                  & 
\multicolumn{1}{c|}{}                & \multicolumn{1}{c|}{}                & \multicolumn{1}{c|}{}                  & 
             & 
\multicolumn{1}{c|}{\ding{51}}                     & 
              
\\ \hline

\citet{mishra2023genpads} & \multicolumn{1}{c|}{}    &  \ding{51}                   
  & 
\multicolumn{1}{c|}{\ding{51}} & 
&   
\multicolumn{1}{c|}{{}}  & 
\multicolumn{1}{c|}{}  &  \ding{51}                       & 
\multicolumn{1}{c|}{\ding{51}}               & 
                  & 
\multicolumn{1}{c|}{}                & \multicolumn{1}{c|}{}                & \multicolumn{1}{c|}{}                  & 
             & 
\multicolumn{1}{c|}{\ding{51}}                     & 
              
\\ \hline

\citet{mishra2023help} & \multicolumn{1}{c|}{}    &  \ding{51}                   
  & 
\multicolumn{1}{c|}{\ding{51}} & 
&   
\multicolumn{1}{c|}{{}}  & 
\multicolumn{1}{c|}{}  &  \ding{51}                       & 
\multicolumn{1}{c|}{}               & \ding{51}
                  & 
\multicolumn{1}{c|}{}                & \multicolumn{1}{c|}{}                & \multicolumn{1}{c|}{}                  & 
             & 
\multicolumn{1}{c|}{\ding{51}}                     & 
              
\\ \hline

\citet{firdaus2023mixing} & \multicolumn{1}{c|}{}    &  \ding{51}                   
  & 
\multicolumn{1}{c|}{\ding{51}} & 
&   
\multicolumn{1}{c|}{{\ding{51} }}  & 
\multicolumn{1}{c|}{}  &                        & 
\multicolumn{1}{c|}{\ding{51}}               & 
                  & 
\multicolumn{1}{c|}{}                & \multicolumn{1}{c|}{}                & \multicolumn{1}{c|}{}                  & 
             & 
\multicolumn{1}{c|}{\ding{51}}                     & 
              
\\ \hline
\citet{mishra2023therapist} & \multicolumn{1}{c|}{}    &  \ding{51}                   
  & 
\multicolumn{1}{c|}{\ding{51}} & 
&   
\multicolumn{1}{c|}{{}}  & 
\multicolumn{1}{c|}{}  &  \ding{51}                       & 
\multicolumn{1}{c|}{}               & \ding{51}
                  & 
\multicolumn{1}{c|}{}                & \multicolumn{1}{c|}{}                & \multicolumn{1}{c|}{}                  & 
             & 
\multicolumn{1}{c|}{\ding{51}}                     & 
              
\\ \hline

\end{tabular}
\end{adjustbox}  
    \label{problemdef_generate}
\end{table}
%%%%%%%%%%%%%%%%%%%%%%%%%%%%%%%%%%%%%%%%%%%%%%%%%%%%%%%%%%%%%%%%%%%%%%%%%%%%%%%%%%%%%%%%%%%%%%%%%%%%

\subsection{Politeness as Natural Language Generation Task}
Several computational politeness research works concentrate on the generation of polite sentences. Given the non-polite text, the objective of the politeness generation task is to generate a politeness-oriented text. For instance, \textit{\enquote{How can we help?}} is a non-polite sentence that can be converted into a polite one: \textit{\enquote{Help has arrived! We are sorry to see that you are having trouble. How can we help?}}. Table \ref{problemdef_generate} provides a brief overiew of the different datasets, languages, approaches, challenges, etc., on computational approaches to politeness generation. 

\citet{madaan2020politeness} focus on politeness as a style transfer task. The authors devise a tag and generate framework for transforming non-polite sentences into polite ones. \citet{fu2020facilitating} propose a computational framework for generating a paraphrased version of a given message that meets the required degree of politeness in a particular communication situation. In recent times, there has been a surge in interest in incorporating politeness besides various styles, emotions, and personalities in conversational artificial intelligent agents to make them behave human-like. The work of \citet{walker1997improvising} represents an initial effort to infuse politeness into conversational agents that provide interesting variations of character and personality in an interactive narrative application.  \citet{gupta2007rude} explore building more affective and socially intelligent task-oriented conversational agents by incorporating different politeness strategies in the culinary instructions. \citet{niu2018polite} attempt to induce politeness in chit-chat conversations in the absence of parallel data. 
\citet{golchha2019courteously} and \citet{firdaus2020incorporating} present a method for inducing polite phrases in customer-care responses to enhance user satisfaction. %by exploiting reinforced pointer networks. 
\citet{wang2020can} infuse the social language, \textit{viz.} politeness and positivity in customer support agents' responses. According to Merriam-Webster dictionary\footnote{\url{https://www.merriam-webster.com/dictionary/positivity}}, positivity is the quality or state of being positive. The authors argue that the users would prefer a conversational agent exhibiting more polite and positive behavior and would be more willing to engage with, respond to, and persist in the interaction when conversing with such agents. Their work is motivated by previous research works that have shown that positivity can engage employees
and improve their performance in the workplace \cite{sweetman2010power} and leads to likeability as it is the language that is upbeat and cheerful \cite{dainton1994maintenance}. \citet {rana2021effect} use the state-of-the-art tag and generate approach to introduce politeness in normal chatbot responses and study the effects of polite triggers among different
genders, age groups, and personalities using a cross-sectional analysis. Lately, \citet{firdaus2022being} also attempt to show the variation in politeness according to the specified personas. Notably, they introduce an approach to generate polite and personalized dialogue responses according to the user’s personal information, primarily based on the age and gender of the user, while being contextually consistent with the conversational history. \citet{saha2022countergedi} propose an ensemble framework for generating polite, detoxified, and emotionally charged counterspeech to combat the rising online hate speech. \citet{silva2022polite} extend the state-of-the-art generative \cite{niu2018polite} and rewriting \cite{madaan2020politeness} approaches to overcome the domain-shift problem and enable the transfer of politeness patterns to a new fashion domain. \citet{sennrich2016controlling} model the politeness interpretation as a machine translation task for English $\rightarrow$ German. %The authors have created parallel training data by labeling the source side with an additional feature that identifies the level of honorific present on the target side. They then use an Attentional Neural Machine Translation (NMT) based system \cite{bahdanau2015neural} to control the honorifics in translation from English $\rightarrow$ German. 
\citet{feely2019controlling} use a similar formulation and controls the honorifics (politeness) in NMT from English $\rightarrow$ Japanese. 

%Include works
Several recent studies have delved into the task of generating polite responses, taking into account a range of affective states and socio-linguistic cues \cite{firdaus2022polise,firdaus2023mixing,mishra2023help,mishra2023pal,ijcai2023p686,mishra2023therapist}. \citet{firdaus2022polise} focus on incorporating user sentiment information to produce contextually appropriate polite responses. Likewise, \citet{firdaus2023mixing} leverage the user's affective states, encompassing emotional and sentiment information, to generate polite responses that exhibit relevant polite behaviors such as greetings, apologies, assurances, appreciations, or expressions of empathy. \citet{mishra2023help} introduce a system for generating polite and empathetic responses, while their subsequent study by \citet{mishra2023pal} utilizes users' emotional information for the same purpose. \citet{ijcai2023p686} introduce a dialogue system capable of generating persuasive responses that adapt politeness and empathy strategies. \citet{mishra2023therapist} incorporate factors such as gender, age, persona, and sentiment information of users to generate contextually relevant polite and interpersonal responses.

\subsection{A note on languages}
% The majority of research in computational politeness focuses on English. However, a few research in other languages has also been reported, \textit{viz.}, Hindi \cite{firdaus2020incorporating,kumar2014developing}, Chinese \cite{li2020studying}, German \cite{sennrich2016controlling}, and Japanese \cite{feely2019controlling}. Recently, \cite{viswanathan2019controlling} developed a domain adaptation technique to bias a model to produce translations according to a desired formality or register. The authors have experimented with formality levels for different language pairs: English $\rightarrow$ French, English $\rightarrow$ Spanish and English $\rightarrow$ Czech.
Politeness is an important facet of communication and is sometimes argued to be cultural-specific, yet existing computational linguistic studies of politeness are primarily dominated by the English language. A few research works in other languages, \textit{viz.}, Hindi \cite{kumar2014developing,firdaus2020incorporating}, German \cite{sennrich2016controlling}, Japanese \cite{feely2019controlling} and Chinese \cite{li2020studying} have also been reported. Similar to politeness research in the English language, there exists a dual focus on the research on politeness analysis in other languages, with some studies concentrating on politeness identification, while others center their efforts on politeness generation. The authors in \cite{kumar2014developing,li2020studying,srinivasan2022tydip} formulate the politeness analysis as an identification task. \citet{kumar2014developing} and \citet{li2020studying} attempt to identify politeness in natural language text, focusing on Hindi and Chinese languages, respectively. \citet{srinivasan2022tydip} evaluate how well multilingual models can identify politeness in text. They conduct this study in nine typologically diverse languages, namely Hindi, Tamil, Korean, Spanish, French, Vietnamese, Russian, Afrikaans, and Hungarian.

Several studies like the works in \cite{sennrich2016controlling,feely2019controlling,viswanathan2019controlling,firdaus2020incorporating} approach politeness from the perspective of a generation task, aiming to generate or maintain politeness in response to given inputs. The authors in \cite{sennrich2016controlling,feely2019controlling,viswanathan2019controlling} devise a methodology to regulate the degree of formality or politeness in the neural machine translation (NMT) from one language to another. \citet{sennrich2016controlling} and \citet{feely2019controlling} perform a pilot study to control honorifics in NMT focusing on English $\rightarrow$ German and English $\rightarrow$ Japanese, respectively. Recently, \citet{viswanathan2019controlling} develop a domain adaptation technique to bias a model to produce translations according to a desired formality or register. The authors experiment with formality levels for different language pairs: English $\rightarrow$ French, English $\rightarrow$ Spanish, and English $\rightarrow$ Czech. \citet{firdaus2020incorporating} propose methods to induce courteous behavior in generic customer-care responses in multilingual scenarios (English and Hindi). 

Some of the aforementioned studies have made significant contributions to dataset creation, while others have contributed significantly to the development of novel approaches and methodologies. The specifics of the datasets and methodologies introduced in these works are elaborated upon in the Dataset and Approaches sections of this manuscript.

%==================================================================================================%

\section{Datasets}
\label{section4}

In this section, we outline the datasets utilized for computational politeness research. We categorize these datasets into non-conversational data (such as request posts on online forums) and conversational data (such as user-agent interaction). %Table \ref{datasets} briefly reports the previous studies using these two types of datasets.%\\

\subsection{Non-conversational Dataset}
%\noindent \textbf{Non-conversational Dataset. }
Social media, websites, blogs, and emails facilitate access to a vast amount of user-generated content in the form of text. These textual data may %contain variety of information 
be of various types like user queries, complaints, etc. All these texts involve making requests in one or another form; requests for seeking information, requests for help, or requests for specific action. It is generally anticipated that such requests will be couched in remarkably polite language to ingratiate the requestor with the person in a position to honor the request. However, due to text length constraints set by some online platforms, this text tends to be brief and lacks the necessary verbal and behavioral signals that help to grasp and interpret a given piece of communication as polite or impolite. Despite such barriers, datasets of requests have been popular for studying politeness from a computational point of view. The reason is that the requests involve the use of B\&L's negative politeness strategy: minimizing the imposition on the addressee, for example, by apologizing for the imposition \enquote{\textit{I'm really sorry but...}} or by being indirect, \enquote{\textit{Would you mind if...}}. 

%The pioneering research by \cite{danescu2013computational} proposes a benchmark corpus called Stanford Politeness Corpus (SPC), consisting of more than 10,000 Wikipedia and Stack Exchange requests. Each request in the dataset is annotated with a discrete score ranging from -3 representing \enquote{very impolite} through 0 indicating \enquote{neutral} to +3 indicating \enquote{very polite} by five Amazon's Mechanical Turk (AMT) workers. The authors eventually assign the politeness score of a request as the average of the five scores assigned by the annotators. 
The pioneering research by \citet{danescu2013computational} proposes a benchmark corpus called Stanford Politeness Corpus (SPC) of requests. The SPC consists of requests from two large online communities: Wikipedia, where the requests involve editing and other administrative functions, and Stack
Exchange (SE), where the requests center around a diverse range of topics (e.g., programming, gardening, cycling). The intuition leading to domain selection is that both communities exhibit a notable abundance of user-to-user requests characterized by substantive interactions rather than superficial social exchanges. These requests typically involve soliciting specific information or concrete actions, underscoring the expectation of a meaningful response from participants. The corpus consists of more than 10,000 requests (4,353 requests from Wikipedia and 6,604 from Stack Exchange). Each request is annotated with a discrete score ranging from -3 representing \enquote{very impolite} through 0 indicating \enquote{neutral} to +3 indicating \enquote{very polite} by five Amazon's Mechanical Turk (AMT) workers. The authors eventually assign the politeness score of a request as the average of the five scores assigned by the annotators. These requests are categorized into 5 positive politeness strategies (\textit{Gratitude, Deference, Greeting, Positive Lexicon, Negative Lexicon} and 15 negative politeness strategies (\textit{Apologizing, Please, Please start, Indirect (btw), Direct question, Direct start, Counterfactual modal, Indicative modal, 1st person start, 1st person pl., 1st person, 2nd person, 2nd person start, Hedges, Factuality}). %An example for each of the labels is provided in the Table \ref{politeness_ex}.

\citet{li2020studying} create a politeness corpus containing 5,300 posts each posted on Twitter and Weibo in 2014. Two native language annotators independently label each post with a politeness score using a similar annotation scheme as in \citet{danescu2013computational}. \citet{madaan2020politeness} introduce a large dataset comprising 1.39 million sentences annotated with politeness scores. These sentences are curated from email exchanges of requests in the Enron corpus \cite{klimt2004introducing}. The sentences are labeled with a politeness score using a politeness classifier \cite{niu2018polite}, and sentences with a score of over 90\% are considered as polite. \citet{chhaya2018frustrated} create a dataset consisting of 1,050 emails extracted from the Enron email dataset and annotate each email for frustration, formality, and politeness labels following the annotation protocol of the Likert Scale. %\cite{allen2007likert}. 
Frustration is labeled on a 3-point scale, with neutral equating to not frustrated; and frustrated and very frustrated are marked with -1 and -2, respectively. Formality and politeness follow a 5-point scale from -2 to +2, where both extremes mark the higher degree of presence and absence of the respective dimension. Each email is assigned a final score as an average of 10 scores annotated by the AMT workers. 

\citet{sennrich2016controlling} utilize OpenSubtitles \cite{tiedemann2012parallel}
and present a politeness annotated corpus for machine translation task from English $\rightarrow$ German. Each sentence in the corpus is automatically annotated with the rules based on a morphosyntactic annotation by \cite{sennrich2013exploiting}. The sentences with imperative verbs are labeled as informal, while sentences containing informal or polite German address pronouns are labeled with the corresponding class. Lately, \citet{bharti2023politepeer} introduce \textit{PolitePEER}, a large-scale corpus of peer reviews annotated with politeness labels. It comprises review sentences from various sources such as ICLR, NeurIPS, Publons, and ShitMyReviewersSay, covering diverse topics including machine learning, natural language processing, artificial intelligence, environment and ecology, clinical medicine, psychiatry and psychology, and engineering. A total of 1348 review sentences from ICLR, 58 from NeurIPS, 888 from ShitMyReviewersSay, and 206 from Publons, are annotated with five politeness labels, namely highly\_impolite, impolite, neutral, polite, and highly\_polite. The authors argue that the \textit{PolitePEER} dataset is one of its kind, which may enable researchers to gain deeper insights into the language used in peer reviews and lead to the development of more accurate models for predicting politeness in this domain. %They observed that the neutral class sentences dominate the sentences belonging to other classes. The authors mentioned that the imbalance in politeness labels can be attributed to the nature of the peer reviews domain. Most reviewers tend to follow a neutral trend while expressing their views. The authors argued that the \textit{PolitePEER} dataset is one of its kind, which may enable researchers to gain deeper insights into the language used in peer reviews and lead to the development of more accurate models for predicting politeness in this domain.} %The authors discerned an inter-annotator agreement of 90.74\%, 90.74\%, and 93.17\% using Cohen’s Kappa \cite{cohen1960coefficient}, Kendall’s Tau \cite{kendall1939problem}, and Krippendorff’s Alpha \cite{ford2004content} measures on a subset of dataset containing 1500 review sentences, which establishes the sufficiently good quality of the dataset.} 

Politeness is a phenomenon that is highly culture and language-specific, and what is polite in one language may be impolite in another, resulting in numerous cross-cultural misunderstandings and conflicts. Thus, it is essential to have a
sound knowledge of politeness across languages and ensure that the machines have the same understanding. Politeness in low-resource Indian languages is a little-explored aspect of the language. Recently, with the rise in online content in
Hindi, a few researchers focus on studying politeness in Hindi texts. \citet{kumar2014developing} introduce a politeness annotated corpus consisting of over 479,000 blog posts and blog comments in Hindi. The blogs are collected using a popular aggregator of blogs in Hindi named \textit{Chitthajagat}. The blogs in the dataset are annotated with one of the four levels of politeness, \textit{viz.} neutral, appropriate, polite, and impolite in a semi-supervised manner. A total of 30,000 texts from the corpus are manually annotated by four annotators. The manually annotated text is utilized for training
classifiers and annotating the remaining blogs in the corpus. \citet{kumar2012challenges} present a Corpus of Computer-Mediated Communication in Hindi (CO3H) for studying politeness in Hindi. The data for the corpus is collected from blogs, web portals, e-forums, emails, and public and private chats over the web. The data consists of comments, including requests, offers, compliments, criticism, etc. After a fundamental analysis of how politeness or impoliteness is expressed in the language, an annotation scheme has been developed for annotating the corpus. Approximately 26,000 comments are manually annotated with one of the four politeness categories, \textit{viz.} neutral, appropriate, polite, and impolite \cite{kumar2021towards}. Table \ref{datasets_nonconv} presents a summary of these non-conversational datasets.

%%%%%%%%%%%%%%%%%%%%%%%%%%%%%%%%%%%%%%%%%%%%%%%%%%%%%%%%%%%%%%%%%%%%%%%%%%%%%%%%%%%%%%%%%%%%%%%%%%%%
\renewcommand{\arraystretch}{1.4}
\begin{table}[htbp]
    \centering
    \caption{Summary of politeness-labeled non-conversational datasets. Details of the evaluation metrics can be found in Section \ref{section6}.}
\begin{adjustbox}{max width = \linewidth}
\begin{tabular}{|l|p{3cm}|p{6cm}|l|p{6cm}|p{3cm}|}
\hline
\multicolumn{1}{|c|}{\textbf{Reference}} & \multicolumn{1}{c|}{\textbf{Dataset}} & \multicolumn{1}{c|}{\textbf{Domain}} & \multicolumn{1}{c|}{\textbf{Language}} & \multicolumn{1}{c|}{\textbf{Task}} & \multicolumn{1}{c|}{\textbf{Evaluation Metrics}}\\ \hline
\citet{danescu2013computational} & Stanford Politeness Corpus of requests & \begin{tabular}[c]{@{}l@{}}Wikipedia and Stack Exchange \\ Request Posts\end{tabular}  & English & Politeness classification & Accuracy \\ \hline
\citet{li2020studying} & Politeness Corpus  & Twitter and Weibo Posts & English and Chinese & Politeness classification & Accuracy, ROC-AUC\\ \hline
{\citet{madaan2020politeness}} & Enron Corpus & Email Exchanges  & English & Style transfer (Non-polite to polite) & BLEU-s \\ \hline
\citet{chhaya2018frustrated} & Enron Dataset & Email  & English & Frustration, formality and politeness classification & Accuracy, Mean Square Error\\ \hline
\citet{sennrich2016controlling} & OpenSubtitles & Movies Subtitles  & English, German & Politeness preservation in neural machine translation & BLEU\\ \hline
\citet{bharti2023politepeer} & PolitePEER & Peer Reviews  & English & Politeness classification & F1-score \\ \hline
\citet{kumar2012challenges} & CO3H & Online Forums Requests & Hindi & Politeness classification & Accuracy \\ \hline
\citet{kumar2014developing} & Politeness Corpus of Hindi Blogs & Blogs Posts and Comments  & Hindi & Politeness classification & Accuracy \\ \hline
\end{tabular}
\end{adjustbox}
\label{datasets_nonconv}
\end{table}
%%%%%%%%%%%%%%%%%%%%%%%%%%%%%%%%%%%%%%%%%%%%%%%%%%%%%%%%%%%%%%%%%%%%%%%%%%%%%%%%%%%%%%%%%%%%%%%%%%%%

% \noindent \textbf{Conversational Dataset. }
%The advancement of artificial intelligence (AI) and natural language processing (NLP) has led to the development of automatic systems that can mimic human characteristics in communication. One of the crucial aspects of human communication is politeness; consequently, several conversational datasets for politeness have been reported in the literature to build end-to-end dialogue systems having a courteous demeanor. 
\subsection{Conversational Dataset}
Politeness has also been explored in the conversational settings. \citet{wang2020can} propose a dataset consisting of 233,571 message pairs exchanged between the ride-sharing company's on-boarding drivers and customer support representative (CSR). Each message pair is annotated automatically using the state-of-the-art pre-trained classifiers with politeness and sentiment scores. \citet{golchha2019courteously} introduce a conversational dataset, CYCCD (Courteously Yours Customer Care Dataset), which consists of the interactions between customers and professional customer care agents of companies on their Twitter handles. The dialogues in the dataset consist of generic utterances in English and their corresponding courteous versions. 

Inspired by their work, \citet{firdaus2020incorporating} follow their guidelines and build a multi-lingual conversational dataset for English and Hindi to generate polite customer care responses in both languages, thereby enhancing the performance and usability of conversational agents. \citet{firdaus2022polise} annotate the CYCCD \cite{golchha2019courteously} with sentiment (positive, neutral, negative) to generate sentiment-guided polite responses. \citet{firdaus2023mixing} annotate the CYCCD \cite{golchha2019courteously} with polite behavior (greeting, apology, assurance, appreciation, and empathy) to generate polite behavior-aware responses in customer support dialogues. \citet{bothe2021polite} explore the DailyDialog \cite{li2017dailydialog} dataset and annotate each utterance of this dataset with a degree of politeness on the scale of [1,5] using a politeness analyzer from the work of \cite{bao2021conversations}. %where a value around 3 indicates neutral, the value inclined towards 1 on the politeness scale indicates impolite, and the value inclined towards 5 on the politeness scale indicates polite. 

\citet{mishra2022please} utilize the recently released multi-domain task-oriented conversational dataset, MultiDoGo \cite{peskov2019multi}, and annotate each utterance of the dataset with one of the four politeness labels, \textit{viz.} polite, somewhat\_polite, somewhat\_impolite, and impolite. The authors perform the annotation in two stages. In the first stage, the politeness score of each utterance of the MultiDoGo dataset is obtained via the Stanford Politeness Classifier (trained on Wikipedia requests data \cite{danescu2013computational}). The utterance is considered polite if the score >= 0.50, else impolite. In the second phase, the authors randomly select approximately 10K utterances from each domain and manually annotate each utterance with one of the four fine-grained politeness classes to obtain the ground-truth data. \citet{mishra2022predicting} follow similar guidelines as in \cite{mishra2022please} and create two politeness-annotated conversational datasets utilizing the DSTC1 \cite{williams2014dialog} and Microsoft Dialogue Challenge (MDC) \cite{li2018microsoft} datasets. \citet{mishra2023genpads} enrich the TaskMaster \cite{byrne2019taskmaster} dataset by manually annotating it with fine-grained politeness labels with the help of crowd-workers from Amazon Mechanical Turk (AMT).

\citet{mishra2022pepds} annotate the PersuasionForGood dataset \cite{wang2019persuasion} with the politeness strategy labels: bald on-record, positive politeness, negative politeness, and off-record. \citet{priya2023multi} curate a large-scale, novel dialogue dataset, POEM, consisting of 5,000 dialogues in English for mental health counseling and legal aid of women and children crime victims. The POEM dataset is manually annotated with politeness labels and multiple emotion labels at the utterance level. For politeness annotation, the authors consider ternary labels, \textit{viz.} polite, neutral, and impolite. For emotion annotation, they consider a rich set of 17 emotion categories, namely anticipation, confident, hopeful, anger, sad, joy, compassion, fear, disgust, annoyed, grateful, impressed, apprehensive, surprised, guilty, trust, and neutral. 

\citet{mishra2023help} also curate a novel mental health and legal counseling dialogue (MHLCD) dataset in English comprising 1006 dialogues between the victim and the counselor. Further, they annotate the counselor's utterances in this dataset with politeness and empathy labels. The authors use ternary politeness labels - polite, neutral, and impolite, and binary empathy labels - empathetic and non-empathetic for politeness and empathy annotations. \citet{mishra2023pal} introduce two large-scale counseling conversational datasets, EPE-enEIH and EPE-HLCC, that focus on supporting individuals dealing with crime victimization and substance addiction. The dataset focused on delivering counseling support to crime victims is prepared by translating the Hindi conversational dataset, EmoInHindi \cite{singh2022emoinhindi} to English. The other dataset used in the study is prepared using High-quality and Low-quality Counseling Conversations data \cite{perez2019makes}, consisting of conversations between counselors and substance addicts. They annotate the agent's utterances in these datasets with one of the 11 emotion categories (caring, hopeful, content, surprised, afraid, confident, proud, trusting, joy, sad and angry), one of three politeness labels (impolite, low\_polite and highly\_polite) and ternary empathy labels (non-empathetic, low\_empathetic and highly\_empathetic).

\citet{ijcai2023p686} curate another counseling conversational dataset, HEAL, consisting of 216 conversations between the counseling bot and the crime victim. The counseling bot's responses are annotated with novel empathy strategies (reflective listening, confidential comforting, evoke motivation, express emotional support, offer counseling, escalate assurance and no strategy) and politeness strategies (positive politeness, negative politeness, and bald-on-record) \cite{brown1978universals}. \citet{mishra2023therapist} introduce a novel psychotherapeutic conversations dataset, \textsc{PsyCon}, comprising of conversations between the therapist and the user suffering from one of the seven most common psychological issues, \textit{viz}. depression, anxiety, stress, bipolar disorder, disruptive behavior and dissocial disorders, post-traumatic stress disorder (PTSD), and schizophrenia \cite{WHO2022mental}. The user's utterances are annotated with sentiment labels (positive, negative, and neutral). The therapist's utterances are annotated with politeness labels (polite, moderately\_polite, and impolite) and interpersonal behavior labels (directing, helpful, understanding, complaint, imposing, confrontational, dissatisfied, and uncertain). Table \ref{datasets_conv} provides a comprehensive summary of these conversational datasets across various dimensions.  

%%%%%%%%%%%%%%%%%%%%%%%%%%%%%%%%%%%%%%%%%%%%%%%%%%%%%%%%%%%%%%%%%%%%%%%%%%%%%%%%%%%%%%%%%%%%%%%%%%%%
\renewcommand{\arraystretch}{1.3}
\begin{table}[htbp]
    \centering
    \caption{Summary of politeness-labeled conversational datasets. Details of the evaluation metrics can be found in Section \ref{section6}.}
\begin{adjustbox}{max width = \linewidth}
\begin{tabular}{|p{4cm}|p{5cm}|p{6cm}|l|p{3cm}|p{3cm}|}
\hline
\multicolumn{1}{|c|}{\textbf{Reference}} & \multicolumn{1}{c|}{\textbf{Dataset}} & \multicolumn{1}{c|}{\textbf{Domain}} & \multicolumn{1}{c|}{\textbf{Language}} & \multicolumn{1}{c|}{\textbf{Task}} & \multicolumn{1}{c|}{\textbf{Evaluation Metrics}} \\ \hline
\citet{golchha2019courteously} & CYCCD & Customer Care  & English & Polite response generation & BLEU, ROUGE-1, ROUGE-2, ROUGE-L, Perplexity, Content Preservation, Emotion Accuracy\\ \hline
\citet{firdaus2020incorporating} & Hindi Customer Care Dialogue & Customer Care  & English, Hindi & Polite response generation & BLEU, ROUGE-1, ROUGE-2, ROUGE-L, Perplexity, Content Preservation, Emotion Accuracy \\ \hline
\citet{wang2020can} & CSR-Driver Exchanges & Customer Care & English & Polite response generation & BLEU, word2vec-based similarity\\ \hline
\citet{bothe2021polite}, \citet{priya2023multi} & Politeness-annotated DailyDialog & Daily-life Conversations  & English & Politeness classification & Accuracy, F1-score\\ \hline
\citet{mishra2022please} & Politeness-annotated MultiDoGO & Airline, Fastfood, Finance, Insurance, Media, Software & English & Polite dialogue generation & Success Rate, Politeness Adaptability, Turn Completion Rate\\ \hline
\citet{mishra2022predicting} & Politeness-annotated DSTC-1, MDC & DSTC-1: Bus Schedule; MDC: Movie-Ticket Booking, Restaurant reservation, Taxi Ordering  & English & Politeness classification & Accuracy, Geometric Mean, Index Balance Accuracy \\ \hline
\citet{mishra2023genpads} & Politeness-annotated TaskMaster & Flight, Food Ordering, Hotel Movie, Music, Restaurant, Sports  & English & Politeness-adaptive dialogue generation & Dialogue Length, METEOR, ROUGE-2-L, Success Rate, Average Politeness Score\\ \hline
\citet{firdaus2022polise} & Sentiment-annotated CYCCD & Customer Care  & English & Sentiment-guided polite response generation & Perplexity, BLEU, Rouge-L, Politeness Adaptability\\ \hline
\citet{mishra2022pepds} & Politeness-annotated PersuasionForGood & Charity Donation  & English & Polite and empathetic persuasive dialogue generation & Perplexity, Response Length, PerStr, PolSt, Emp\\ \hline
\citet{priya2023multi} & POEM & Mental Health and Legal Counseling & English & Politeness and Emotion Classification & Accuracy and F1\\ \hline
\citet{mishra2023help} & MHLCD & Mental Health and Legal Counseling & English & Polite and empathetic dialogue generation & Perplexity, R-LEN, CoStr, Pol, Emp \\ \hline
\citet{mishra2023pal} & EPE-enEIH and EPE-HLCC & Mental Health and Legal Counseling & English & Emotion-adaptive polite and empathetic dialogue generation & Perplexity, Response Length, EPC, EEC, PC, EC\\ \hline
\citet{ijcai2023p686} & HEAL & Mental Health and Legal Counseling  & English & Politeness and empathy strategies- adaptive persuasive dialogue generation & Perplexity, Response Length, CoAct, EmpStr, PolStr  \\ \hline
\citet{firdaus2023mixing} & Polite Behavior-annotated CYCCD & Customer Care & English & Polite behavior-aware response generation  & Perplexity, BLEU, ROUGE-L, Politeness Adaptability\\ \hline
\citet{mishra2023therapist} & PSYCON & Mental Health Counseling & English & Polite and interpersonal behavior-aware response generation & Perplexity, BERTScore-F1, Response Length, GA$_c$, P$_c$, CT$_c$, Po$_c$, IB$_c$ \\ \hline
\end{tabular}
\end{adjustbox}
\label{datasets_conv}
\end{table}
%%%%%%%%%%%%%%%%%%%%%%%%%%%%%%%%%%%%%%%%%%%%%%%%%%%%%%%%%%%%%%%%%%%%%%%%%%%%%%%%%%%%%%%%%%%%%%%%%%%%
%==================================================================================================%

\section{Approaches}
\label{section5}

In this section, we present the approaches used for computational politeness. In general, approaches to computational politeness can be categorized into statistical and deep learning-based approaches. We will examine these methods in the following sections.

\subsection{Statistical Approaches}
This section describes the feature sets and the the learning algorithms that have been reported for statistical computational politeness. The majority of the approaches use bag-of-words as features. Nevertheless, several other linguistic features have been reported in addition to these features. Table \ref{features} summarizes the features used in past work.

%%%%%%%%%%%%%%%%%%%%%%%%%%%%%%%%%%%%%%%%%%%%%%%%%%%%%%%%%%%%%%%%%%%%%%%%%%%%%%%%%%%%%%%%%%%%%%%%%%%%
\begin{table}[hbt!]\scriptsize
    \centering
    \caption{Features used for statistical classifiers. (\# N) represents the number of respective features used in the experiment.}
\begin{adjustbox}{max width=\linewidth}
    \begin{tabular}{|c|l|}
\hline
                                \textbf{Reference}                 & \multicolumn{1}{c|}{\textbf{Features}}                         \\ \hline
 {\citet{danescu2013computational}} & 
Lexical and syntactic features (\# 20) \\ \hline
 {\citet{li2020studying}}           & PoliteLex (\# 26), LIWC, EmoLex, Stanford Politeness API               \\ \hline
 {\citet{kumar2021towards}}         & Unigrams, Bigrams and Linguistic features (\# 8)                     \\ \hline
 {\citet{chhaya2018frustrated}} & Lexical (\# 12), Syntactic (\# 6), Derived (\# 8), Affect-based (\# 29), Word embeddings (Word2Vec) \\ \hline
 {\citet{madaan2020politeness}} & N-grams tf-idf \\ \hline
 {\citet{fu2020facilitating}} & Lexical and syntactic features (\# 18) \\ \hline
\end{tabular}
\end{adjustbox}
    \label{features}
\end{table}
%%%%%%%%%%%%%%%%%%%%%%%%%%%%%%%%%%%%%%%%%%%%%%%%%%%%%%%%%%%%%%%%%%%%%%%%%%%%%%%%%%%%%%%%%%%%%%%%%%%%

\noindent \textbf{Approaches to Politeness Identification.} \citet{alexandrov2007constructing} present numerical estimations for politeness based on the lexical-grammatical properties of the text and subjective expert opinions. They have considered three factors of politeness: the first
greeting, polite words, and polite grammar forms, and used these factors to construct a series of polynomial models for automatic evaluation of the level of politeness. \citet{alexandrov2008regression} propose a linear regression model for politeness estimation using the same approach as in \cite{alexandrov2007constructing}.

%Most work in statistical computational politeness relies on Support Vector Machines (SVM). 
In the past, several works in statistical computational politeness have been reported that rely on Support Vector Machines (SVM). \citet{danescu2013computational} construct a classifier for predicting politeness in natural language requests in English with various domain-independent lexical and syntactic features about key politeness theories, for instance, a first or second person start vs. plural, please start vs. please, etc. Specifically, they propose two classifiers - a bag of words classifier (BOW) and a linguistically informed classifier (Ling.). BOW classifier is an SVM with unigram feature representation. The Ling. classifier is an SVM using the linguistic features\footnote{A total of 20 politeness features have been identified by the authors, namely \textit{Gratitude}, \textit{Deference},\textit{Counterfactual modal}, to name a few.} besides the unigram features. They then apply the logistic regression model on top of the SVM output to predict the politeness scores for requests between 0 and 1. However, \citet{hoffman2017evaluating} challenge the work of \citet{danescu2013computational} and revalidates the politeness labeling tool for broadly addressing politeness in social computing data. 

\citet{li2020studying} develop a politeness feature set, \textit{PoliteLex}, that is compatible with both English and Mandarin to study the resemblances and distinctions between the US and China's politeness. They have trained several SVM classifiers using features from one or more feature sets, \textit{viz.} \textit{PoliteLex}, \textit{Linguistic Inquiry and Word Count (LIWC)}\footnote{\textit{LIWC} maps tokens into psychological categories.} \cite{pennebaker2001linguistic}, \textit{EmoLex}\footnote{\textit{EmoLex} consists of 14,182 English
unigrams mapped to 7 emotion categories – joy, surprise, anticipation, sadness, fear, anger, and
disgust.} \cite{mohammad2013crowdsourcing} and \textit{Stanford Politeness API} \cite{danescu2013computational}. The authors have shown that \textit{PoliteLex}, combined with \textit{LIWC} and \textit{EmoLex} outperform the \textit{Stanford Politeness API} in predicting politeness on Twitter, Weibo, and Stanford Politeness Corpus. Further, the findings of this research reveal that within the Mandarin Weibo platform, discussions centered around future-oriented themes, expressions of group affiliation, and expressions of gratitude are generally regarded as more polite in comparison to English Twitter. Conversely, discussions involving topics related to death, the use of pronouns (excluding honorifics), and informal language tend to exhibit higher degrees of impoliteness on Mandarin Weibo compared to English Twitter.

\citet{kumar2021towards} propose an SVM-based classifier for identifying linguistic politeness in Hindi. The authors have proposed two classifiers based on the bag-of-words (BoW) model. Of these two classifiers, one employs unigram feature representation, and the other employs unigram and bigram feature representations. They have also developed one more classifier which utilizes unigrams, bigrams, and manually identified linguistic structures as features such as formulaic expressions (\textit{dhanyawad, kripya, etc.}), use of particle `\textbf{ji}', `\textbf{zara}', subjunctive verb form (\textit{mujhe, samjhe, etc.}), suggestion markers/deontics (\textit{cahiye}, etc.), honorific pronominals and verb form (\textit{bataiega}, etc.).

\citet{wang2020can} utilize statistical models to estimate the role of politeness and positivity in predicting drivers engagement. They have analyzed that use of such social language leads to greater users’ responsiveness and task completion. In particular, they have employed pre-trained SVM-based classifier \cite{danescu2013computational} and a rule-based sentiment analyzer, VADER \cite{hutto2014vader} to predict the politeness level and positivity of messages, respectively. 

\citet{miyamoto2020politeness} propose an autonomous politeness control method for conversational agents. The authors have constructed a regression model to predict politeness and comfortableness in terms of three social factors of the politeness theory: social distance, social status difference, and conversation scene. They have conducted multiple regression analyses to obtain a model that facilitates the selection of appropriate politeness of utterance according to three social factors. \citet{chhaya2018frustrated} model frustration along with its two complementary affects, formality and politeness. The authors have proposed state-of-the-art statistical models for modeling frustration, formality, and politeness in their work. Specifically, they have developed Linear, Lasso, Ridge, and SVR regression models and different classification models, \textit{viz.} Logistic Regression, SVC, Linear SVC, Random Forest, and Nearest Neighbors, with different features for determining the existence or non-existence of frustration, formality, and politeness in emails. 

\noindent \textbf{Approaches to Politeness Generation.} \citet{madaan2020politeness} introduce a pipeline approach, \textit{tag and generate}, which is a two-staged approach for converting non-polite sentences into polite ones while preserving the content of the source text. The tagger model first identifies the style attribute markers from the source style and then either substitutes them with a positional token named [TAG] or adds [TAG] tokens to the input without deleting any phrase from the input. This feature of the model allows generation of these tags in an input that lacks an attribute marker. The generator model generates sentences in the target style by replacing these [TAG] tokens with stylistically relevant words inferred from the target style. In order to estimate the style markers for a specific style, a simple approach based on n-gram tf-idfs has been used. \citet{fu2020facilitating} propose a pipeline approach for generating a paraphrase of a given message that can convey the desired degree of politeness in a particular communication setting. This pipeline approach consists of three steps: \textit{Plan}, \textit{Delete}, and \textit{Generate}. For a given input message, this approach initially determines the politeness strategies and their respective markers. The proposed model employs Integer Linear Programming (ILP) in the plan stage to estimate an appropriate target strategy combination, followed by the deletion of markers associated with the strategies that need to be omitted to obtain the post-deletion context. The proposed model then sequentially integrates the new strategies from the ILP solution into this context to construct the final paraphrased message. 

\subsection{Deep Learning-based Approaches}

With the gain in popularity of deep learning-based architectures in NLP applications, many such approaches have been reported for computational politeness. 

\noindent \textbf{Approaches to Politeness Identification.} \citet{aubakirova2016interpreting} propose a Convolutional Neural Network (CNN)-based architecture using the Stanford Politeness Corpus for predicting politeness in natural language requests. Furthermore, the authors have presented various visualization strategies: activation clustering, embedding space transformations, and first derivative saliency to understand what and how machines learn about politeness. The authors state that this approach not only outperforms the prior work of \citet{danescu2013computational} using manually-defined features but also extends the existing features and identifies novel politeness strategies such as indefinite pronouns (\textit{something/somebody}) and punctuation (question marks \enquote{\textit{???}}, an ellipsis \enquote{...}). \citet{mishra2022predicting} employ a hierarchical transformer network for predicting the politeness label of a given utterance in a goal-oriented dialogue. The authors state that this approach allows them to learn contextual information from previous utterances, which aids in predicting the politeness labels accurately. \citet{dasgupta2023graph} propose a graph-induced transformer network (GiTN) to automatically detect formality and politeness in text. GiTN utilizes Graph Convolution Network (GCN) \cite{kipf2016semi} to encode the syntactic information present in the input text and the BERT \cite{devlin2018bert} model to generate the contextual embeddings of the input text. The output of GCN and BERT are then combined and passed through an attention-based BiLSTM \cite{yu2019review} layer followed by the fully connected layer to automatically detect stylistic characteristics like formality and politeness in the input text. The authors independently fine-tune GiTN on formality-annotated and politeness-annotated corpora for formality and politeness detection tasks, respectively.

\noindent \textbf{Approaches to Politeness Generation.} \citet{sennrich2016controlling} introduce an attention-based encoder-decoder NMT system \cite{bahdanau2015neural} for English $\rightarrow$ German translation task. The authors demonstrate that using the target side T-V annotation as an extra input during training of an NMT model can control the generation of politeness (honorifics) at test time via side constraints. \citet{feely2019controlling} propose a formality-aware NMT system for controlling honorifics in English $\rightarrow$ Japanese translation and demonstrate that this approach improves the translation quality, especially for formal and polite sentences in each test set. 

\citet{niu2018polite} propose three weakly supervised models, namely the Fusion model, Label-Fine-Tuning (LFT) model, and Polite Reinforcement Learning (Polite-RL) model. These models aim to generate natural, varied, and contextually consistent polite or impolite responses for open-domain dialogues without parallel data. Each of
the three models is based on a state-of-the-art politeness classifier \cite{danescu2013computational}, and a sequence-to-sequence dialogue model. The Fusion model combines the decoder of an encoder-attention-decoder conversation model with a language model trained on stand-alone polite utterances. The LFT model prepends a politeness-score scaled label to each source sequence during training. The model can generate polite, neutral, and rude responses at test time by simply scaling the label embedding by the corresponding score. The Polite-RL facilitates the generation of polite responses by assigning rewards proportional to the politeness classifier score of the sampled response. The authors have illustrated that the LFT and Polite-RL models can generate better polite responses without compromising the dialogue quality. 

\citet{golchha2019courteously,firdaus2020incorporating,wang2020can} provide evidence that incorporation of courteousness is essential for achieving user satisfaction and promoting user retention. Thus, \citet{golchha2019courteously} propose a reinforced pointer-generator model for generating emotionally and contextually consistent courteous responses in task-oriented customer-care dialogues in English. Motivated by their work, \citet{firdaus2020incorporating} utilizes a pointer-generator network for generating courteous responses in both Hindi and English by modeling the user's emotional state by learning language invariant representation using adversarial training. \citet{wang2020can} propose a sequence-to-sequence learning framework to infuse politeness and positivity in goal-oriented customer support agents' responses while preserving content. \citet{mishra2022please} introduce a novel end-to-end reinforcement learning (RL) based Politeness Adaptive Dialogue System (PADS). This approach first extracts the politeness semantics of an utterance using a politeness classifier based on DistilBERT \cite{sanh2019distilbert}. Then, the RL-based dialogue agent uses this politeness information to design its reward feedback. The authors state that polite rewards in PADS consider the user's satisfaction information, enabling the agent to accomplish the task quickly. 

Lately, \citet{silva2022polite} examine two different strategies for generating politeness, \textit{viz.} politeness response generation \cite{niu2018polite} and rewriting \cite{madaan2020politeness}. They further analyze how each method handles domain changes and adapts each of them to transfer politeness patterns to the novel fashion domain. Their work is one of the earliest studies on politeness for task-oriented dialog agents that facilitated transfer learning for particular domains. \citet{saha2022countergedi} propose \textit{CounterGeDi}, an ensemble of generative discriminators (\textit{GeDi}) to guide the generation of a DialoGPT model toward more polite, detoxified, and emotionally charged counterspeech in order to combat the increasing online hate speech.  \citet{firdaus2022being} propose a Polite Personalized Dialog Generation (PoPe-DG) framework for inducing politeness in the responses of dialogue systems in accordance with the user's persona information such as age and gender. The authors have manually created polite templates according to the user profiles utilizing the bAbI dataset \cite{joshi2017personalization}. Their proposed PoPe-DG framework employed a reinforced deliberation network for encoding the age and gender information along with polite templates to generate polite personalized responses. 

\citet{mukherjee2023polite} introduce a polite chatbot that can generate polite and contextually coherent responses. The authors formulate a three-step method for building a polite chatbot. In the first step, the authors train a BART \cite{lewis2019bart} based politeness transfer model (PTM) in a supervised fashion that generates polite synthetic dialogue pairs of contexts and polite utterances. PTM takes neutral sentences as input and outputs a sentence that preserves the content while increasing politeness. The synthetic input-output pairs are obtained following \cite{madaan2020politeness}. Then, they apply this politeness transfer model to generate synthetic polite chat data, which they finally utilize to train a dialogue model capable of generating polite responses. For the training of the dialogue model, multiple pre-trained language models are employed, namely GPT-2 \cite{radford2019language}, DialoGPT \cite{zhang2019dialogpt}, and BlenderBot \cite{roller2020recipes}.

\citet{firdaus2022polise} introduce an end-to-end Transformer Encoder-Decoder \cite{vaswani2017attention} based architecture to predict the sentiment of the utterances and then utilize the predicted sentiment information to generate contextually correct polite responses in customer support dialogues. In particular, they utilize a hierarchical network consisting of two transformer-based encoders, namely \textit{sentence encoder} (SE) to encode the utterances and \textit{context encoder} (CE) to encode the output of the sentence encoder to capture the dialog context. The authors apply softmax activation on the output of the SE to predict the sentiment information, which is then combined with the contextual information obtained from CE to generate polite responses aligned with the user's sentiment and dialogue history. To ensure that the users’ sentiment and politeness are induced appropriately in the generated responses, the authors design task-specific rewards and jointly employ reinforcement learning (RL) and machine learning to train the proposed model. \citet{firdaus2023mixing} present a transformer-based encoder-decoder network with an affective tracker to capture the contextual information along with the affective information, behavior-aware generators to attend the context information
to compute behavior-aware polite representations, and a polite generator to generate the
final polite response considering the representations from different generators. This framework aims to capture the emotional aspect from the context, generate response representations based on different polite behaviors (greeting, apology, assurance, empathy, and appreciation), and ultimately produce appropriate polite responses utilizing a weighted-sum approach.

\citet{mishra2023genpads} propose GenPADS, a novel end-to-end dialogue system that reinforces politeness in task-oriented conversations. GenPADS incorporates a politeness classifier and a generation model in a reinforcement learning setting to adapt and generate polite responses that are informative, empathetic, and interactive. To build the politeness classifier and generation model, they employ DistilBERT \cite{sanh2019distilbert} and BART \cite{lewis2019bart}, respectively, as the backbone architectures. The development of GenPADS consists of three steps: GenPADS first utilizes a transformer-based politeness classifier to extract the politeness semantics from the utterances. Subsequently, a dialogue agent, operating on a reinforcement learning framework, utilizes this extracted politeness information to formulate its reward feedback mechanism, thereby adapting its behavior to prioritize polite actions. Finally, a transformer-based generator model is employed to generate interactive and diverse responses.

\subsection{Miscellaneous}
In addition to the two approaches mentioned above, a few other approaches for politeness study in NLP have been reported. In this section, we will describe a few such approaches.

\citet{johnson2004politeness} attempt to teach polite behaviors to an animated pedagogical agent to build a socially intelligent system that can monitor learner performance and provide socially sensitive feedback at appropriate times. They have utilized the natural language generator \cite{johnson2004generating} to build a computational framework for producing appropriate interaction tactics. The authors claim that their generation framework could present the same tutorial comment
with different degrees of politeness. For example, a suggestion to save the current factory description can be stated either as bald on record (e.g., \textit{\enquote{Save the factory now}}), as a hint (\textit{\enquote{Do you want to save the factory now?}}), or as a suggestion of what the tutor would do (\textit{\enquote{I would save the factory now}}) or as a suggestion of joint action (\textit{\enquote{Why don’t we save our factory now?}}). \citet{gupta2007rude} present POLLy (Politeness for Language Learning), which consists of two components: an AI Planner based on GraphPlan \cite{blum1997fast} and a Spoken Language Generator (SLG). AI Planner generates a sequence of actions (plan) for completing the cooking task, and the SLG communicates the plan needed to accomplish that task collaboratively, with the ultimate objective of learning English as a second language.

\citet{miller2008computational,miller2007computational,miller2007computational} describe a computational model of etiquette and politeness perception across different cultures. They have designed and implemented a computational algorithm called \enquote{Etiquette Engine (EE)} based on B\&L's politeness theory to serve as a predictive model of an
observer’s perceived \enquote{degree of imbalance} of an interaction, whether among humans or humans and non-player characters (NPCs). They have demonstrated EE's ability to produce culture-specific, politeness-appropriate utterances and perceptions of utterances in controlled tests involving project team members, open surveys, and game settings.

\citet{roman2004politeness,roman2006politeness} present empirical pilot studies on the role of politeness in dialogue summarization. The authors have created four dyadic dialogues between a customer and vendor using the NECA\footnote{NECA is a conversational agent platform that allows to control several dialogue parameters such as interlocutors' role, interests, and politeness.} (\textit{Net Environment for Embodied Emotional Conversational Agents}) system. They further asked thirty participants of different qualifications, genders, and courses involved in the study to summarize these dialogues. %The authors have used four dialogues automatically generated by the Net Environment for Embodied Conversational Agents (NECA) system  and got them summarized by thirty participants of different qualifications, gender, and course. 
These studies demonstrate how people deal with politeness/behavior in dialogues and how their point of view influences their summaries to facilitate the construction of an automatic dialogue summarization system. In \citet{kumar2010translating}, the authors present a corpus-based study of politeness across two languages. The authors have studied politeness in a translated parallel corpus of Hindi and English and showed how politeness in a Hindi text is translated into English. The authors assert that their study can be helpful for machine translation tasks from Hindi $\rightarrow$ English while maintaining the degree of politeness across languages. 

In \citet{song2023appreciation}, based on the B\&L politeness theory \cite{brown1978universals}, the authors examine the effect of e-commerce chatbot politeness strategies (apology versus appreciation) on customers' post-recovery satisfaction, taking into consideration the variable of face concern and further investigated the moderating influence of time pressure on it. Prior investigations within the domain of politeness theory in AI have predominantly centered on customer preferences for polite chatbots and how impolite chatbots can trigger face-threatening acts among customers \cite{dippold2020turn}. Their study extends this line of research by demonstrating that chatbots can employ politeness strategies in response to customer's face-threatening acts resulting from service failures. Their findings suggest that, in post-recovery consumer satisfaction, fostering a positive human-chatbot relationship (through appreciation) proves more effective than acknowledging the chatbot's limited competence (via apology). This insight serves as a valuable reference point for e-commerce platforms seeking to devise suitable service recovery strategies for their chatbots.

\subsection{Politeness and Large Language Models}
 {Large Language Models (LLMs), (a.k.a generative AI), are language models with tens or hundreds of billions (or more) of parameters trained on massive textual data \cite{zhao2023survey}\footnote{In the existing literature, there exists no established agreement regarding the minimal parameter scale deemed necessary for LLMs, primarily due to the interplay between model capacity, data magnitude, and computational resources. In this survey, we adopt a more flexible interpretation of LLMs and prioritize discussions revolving around politeness in LLMs.}. Examples of LLMs include GPT-3 (175B parameters) \cite{brown2020language}, LLaMA2 (13B, 70B parameters) \cite{touvron2023llama}, ChatGPT\footnote{\href{https://chat.openai.com/}{https://chat.openai.com/}}, etc.} %In recent times, there has been a significant surge in the prominence of LLMs. With the release of ChatGPT, they have gained widespread recognition and use across various domains, including research and industry; perhaps most notably, they have become more accessible and well-known to the general public. These unsupervised autoregressive models are capable of forecasting the next tokens, whether they are characters, words, or strings, by leveraging contextual information from previous data. To enable the LLMs to demonstrate their abilities, sophisticated prompt engineering \cite{neuralmagic2023} is required. Prompts are used to probe the LLMs to generate the target outcome by sampling the language distribution. 
The rapid advancements and impressive achievements of LLMs have triggered a revolutionary shift in the NLP tasks like emotion recognition \cite{zhao2023chatgpt}, summarization \cite{pu2023chatgpt}, dialogue \cite{lin2023llm}, etc. %These models have showcased substantial improvements across a variety of traditional tasks in NLP, including natural language understanding tasks like emotion recognition \cite{zhang2023refashioning,zhao2023chatgpt}, hate speech detection \cite{huang2023chatgpt,oliveira2023good}, to name a few, to generative tasks such as summarization \cite{liang2022holistic,pu2023chatgpt}, dialogue \cite{lin2023llm}, code generation \cite{liu2023your}, and more. 
With the ability to generate creative and human-like text that involves knowledge utilization and complex reasoning, whether the LLMs' responses align with politeness norms or values remains a critical question. Since politeness norms substantially affect human-human interactions, several researchers are actively investigating whether the LLMs possess social competence encompassing politeness recognition, interpretation, and understanding, which is necessary for effective communication and social interactions. 

Recently, \citet{li2023well} evaluate the performance of state-of-the-art language models on politeness prediction using the \textit{Stanford Politeness Corpus} (SPC) \cite{danescu2013computational}. %They have fine-tuned BERT on SPC and then used the fine-tuned model for politeness prediction.
They have evaluated the ChatGPT competency in predicting politeness. This study has shown that %language models can be fine-tuned to achieve human-level performance. Further, 
zero-shot prediction with ChatGPT can provide reasonable results and explanations for its prediction. \citet{ziems2023can} prompt the LLMs, namely Flan-T5 \cite{chung2022scaling}, OpenAI's GPT-3 (text-001, text-002, and text-003) \cite{brown2020language} and ChatGPT in a zero-shot fashion for classifying the utterances in SPC into one of three politeness labels, \textit{viz.} polite, neutral, and impolite. %An example of prompting ChatGPT for politeness classification is shown in the box. %Table \ref{tab:prompt} 
They have mentioned that these LLMs performed fairly well on the politeness classification task, which indicates the LLMs' broader ability to recognize conversational social norms. %Specifically, they found that these LLMs performed best at recognizing polite and neutral utterances and worst at identifying impolite utterances. All three models suffered from making accurate judgments towards direct or indirect questions with 1st and 2nd person mentions \cite{danescu2013computational}.

The widespread application of LLMs in various fields, such as healthcare, education, etc. underscores the importance of assessing their adherence to social norms, particularly politeness. This raises the question of whether LLMs should exhibit politeness towards humans and vice versa. \citet{lievin2022can} highlight the significant impact of politeness on human-machine interactions, indicating that adopting polite strategies by LLMs fosters trust, especially in contexts like healthcare and education. Since LLMs like ChatGPT learn from human interactions, behaving politely with them facilitates the generation of respectful and collaborative outputs. While human politeness may not affect LLM behavior, these systems typically refrain from responding to derogatory or offensive language. For example, Microsoft's GPT-powered Bing AI responds to such language by expressing disapproval, while ChatGPT advises against using offensive language or clarifies its lack of emotions.

\section{Reported Performance}
\label{section6}

Table \ref{tab:results_detect} and Table \ref{tab:results_generate} illustrate the reported values from previous research works for politeness identification and politeness generation tasks, respectively. It is important to note that the values cannot be directly compared since they are deduced from distinct datasets, experiment settings, methodologies, or metrics. Besides, in a few earlier studies, simply an analysis of datasets is conducted. These works do not report any evaluation metric such as Accuracy or F1-score, etc. Consequently, we have not included such works in the tables. Nonetheless, the table provides an approximation of the present performance of computational approaches to politeness identification and generation.  

\textbf{Reported Performance on Politeness Identification.} In Table \ref{tab:results_detect}, we have highlighted the performance evaluation for the politeness identification task reported in the literature on the different datasets in varied domains. \citet{danescu2013computational} show that linguistically-informed features outperform the bag-of-words model in both in-domain and cross-domain settings. The authors compare the accuracy of the politeness classifier for both in-domain and cross-domain settings with the human ability to detect politeness and find that their classifier achieves near-human performance in detecting politeness. \citet{li2020studying} report lexical features as their top discriminating features. The authors show that the lexical approach to predicting politeness is generalizable across domains. The CNN-based classifier in \citet{aubakirova2016interpreting} results in the accuracy of 85.8\% and 66.4\% on Wikipedia and Stack Exchange (SE) corpora, respectively. 

\citet{mishra2022predicting} report metrics other than accuracy for evaluating their proposed hierarchical transformer network for politeness prediction. Specifically, they have used F1-score, geometric mean(GM) \cite{barandela2003strategies}, and index-balance accuracy (IBA) \cite{garcia2012effectiveness} as the automatic evaluation metrics. The graph-induced transformer-based architecture in \citet{dasgupta2023graph} reports the F1-scores of 75\%, 75\%, and 81\% for politeness identification on Wikipedia, Stack Exchange, and Enron Corpus. Likewise, \citet{priya2023multi} evaluate the performance of the politeness detection task in dialogues using accuracy and F1-scores. The authors have reported the accuracy and F1 scores of 90.30\% and 87.19\%, and 86.78\% and 75.27\% for politeness identification on the  POEM and DailyDialog datasets, respectively. For the politeness identification task, it has been observed that accuracy and F1 scores stand out as the predominant evaluation metrics.

%%%%%%%%%%%%%%%%%%%%%%%%%%%%%%%%%%%%%%%%%%%%%%%%%%%%%%%%%%%%%%%%

%\renewcommand\arraystretch{1.3}
%\begin{table*}[]\Huge
\begin{table*}[hbt!]\scriptsize
% \caption{Reported values for computational approaches to politeness identification. All performance values are indicated in percentages.}
\caption{Reported performance values (in \%) for computational approaches to politeness identification.}

  \label{tab:results_detect}
  \begin{adjustbox}{max width=\linewidth}
   \begin{tabular}{|l|lll|l|}
\hline
\multicolumn{1}{|c|}{\multirow{2}{*}{\textbf{Reference}}} & \multicolumn{3}{c|}{\textbf{Details}}                                                                                                                                                                   & \multicolumn{1}{c|}{\multirow{2}{*}{\textbf{Reported Performance}}} \\ \cline{2-4}
\multicolumn{1}{|c|}{}                                    & \multicolumn{1}{c|}{\textbf{Dataset}}                                                       & \multicolumn{1}{c|}{\textbf{Language}}  & \multicolumn{1}{c|}{\textbf{Domain}}                            & \multicolumn{1}{c|}{}                                               \\ \hline
 {\citet{danescu2013computational}}                                                      & \multicolumn{1}{l|}{Stanford Politeness Corpus of requests}                                                                            & \multicolumn{1}{l|}{English}            & \begin{tabular}[c]{@{}l@{}}Online Communities -\\ Wikipedia and Stack Exchange (SE)\end{tabular}                &   \begin{tabular}[c]{@{}l@{}}In Domain - \\Acc:  83.79 (Wiki) and 78.19 (SE)\\Cross Domain - \\Acc: 67.53 (Wiki) and 75.43  {(SE)} \end{tabular}                                                                \\ \hline

 {\citet{li2020studying}}                                                        & \multicolumn{1}{l|}{\begin{tabular}[c]{@{}l@{}}Twitter and Weibo posts, \\ Stanford Politeness Corpus of requests\end{tabular}}        & \multicolumn{1}{l|}{\begin{tabular}[c]{@{}l@{}}Chinese and \\English\end{tabular}} & \begin{tabular}[c]{@{}l@{}}Online platforms (Twitter, Weibo, \\ Wikipedia and Stack Exchange)\end{tabular} &  \begin{tabular}[c]{@{}l@{}}Acc: 77.1, F1: 77.2, ROC-AUC: 77.5 (Twitter)\\ Acc: 80.2, F1: 80.4, ROC-AUC: 81.5 (Weibo)\\ Acc: 68.6 ,F1: 68.5, ROC-AUC: 68.4 (SPC)\end{tabular}                                                                      \\ \hline

 {\citet{kumar2021towards}}                                                        & \multicolumn{1}{l|}{CO3H}                                                                                                             & \multicolumn{1}{l|}{Hindi}              & Blog comments                                                                                              &  Acc: 77.55                                                                   \\ \hline

 {\citet{chhaya2018frustrated}}                                                       & \multicolumn{1}{l|}{Enron Dataset}                                                                                                     & \multicolumn{1}{l|}{English}            & Emails                                                                                                     &  MSE: 1.556, Acc: 72             \\ \hline

 {\citet{aubakirova2016interpreting}}                                                      & \multicolumn{1}{l|}{Stanford Politeness Corpus of requests}                                                                            & \multicolumn{1}{l|}{English}            & \begin{tabular}[c]{@{}l@{}}Online Communities - \\ Wikipedia and Stack Exchange\end{tabular}               &    Acc: 85.8 (Wikipedia) and 66.4 (SE)                                                                  \\ \hline

 {\citet{mishra2022predicting}}                                                        & \multicolumn{1}{l|}{\begin{tabular}[c]{@{}l@{}}Politeness-annotated DSTC1  \\ Politeness-annotated Microsoft Dialog Challenge - \\ A Dialogue State Tracking (MDC)\end{tabular}} & \multicolumn{1}{l|}{English}            & \begin{tabular}[c]{@{}l@{}}MDC - Movie, Restaurant, Taxi \\ DSTC1-Bus schedule\end{tabular}                &  \begin{tabular}[c]{@{}l@{}}In Domain - \\ Acc: 98.1, F1:98.5, GM: 98.9, IBA:95.0\\ Cross Domain -\\ Acc: 65.8, F1: 64.0, GM: 72.5, IBA: 53.1\end{tabular}                                                                                  \\ \hline

 {\citet{priya2023multi} }                                                    & \multicolumn{1}{l|}{\begin{tabular}[c]{@{}l@{}}POEM\\Politeness-annotated DailyDialog\end{tabular} }                                                                                                               & \multicolumn{1}{l|}{English}            & Mental health and legal counseling                                                                         &  {\begin{tabular}[c]{@{}l@{}} {Acc: 90.30, F1: 87.19 (POEM)}\\  {Acc: 86.78, F1: 75.27 (DailyDialog)}	\end{tabular} }                                                                   \\ \hline

 {\citet{bharti2023politepeer} }                                               & \multicolumn{1}{l|}{PolitePEER}                                                                                                        & \multicolumn{1}{l|}{English}            & Peer reviews                                                                                               &  \multicolumn{1}{l|}{\begin{tabular}[c]{@{}l@{}}Class-wise F1 scores- \\ highly polite: 98, polite: 89,\\neutral:81, impolite:78,\\highly impolite:92 \end{tabular} }                                                                  \\ \hline

 {\citet{dasgupta2023graph}}                                                      & \multicolumn{1}{l|}{\begin{tabular}[c]{@{}l@{}}Stanford Politeness Corpus of requests \\ Enron Corpus\end{tabular}}                    & \multicolumn{1}{l|}{English}            & \begin{tabular}[c]{@{}l@{}}Online Communities - \\ Wikipedia and Stack Exchange\\ Emails\end{tabular}      &  \begin{tabular}[c]{@{}l@{}} {F1 Scores -}\\
 {75 (Wikipedia), 75 (SE), 81 (Enron)}\end{tabular}                                                                   \\ \hline

\end{tabular}
\end{adjustbox}
\end{table*}

%%%%%%%%%%%%%%%%%%%%%%%%%%%%%%%%%%%%%%%%%%%%%%%%%%%%%%%%%%%%%%%%%%%%%%%%%%%%%%%%

%%%%%%%%%%%%%%%%%%%%%%%%%%%%%%%%%%%%%%%%%%%%%%%%%%%%%%%%%%%%%%%%

\begin{table*}[hbt!]\scriptsize
% \caption{Reported values for computational approaches to politeness generation. All performance values are indicated in percentages except BLEU and PPL.}
\caption{Reported performance values (in \% except BLEU and PPL) for computational approaches to politeness generation.}

  \label{tab:results_generate}

  \begin{adjustbox}{max width=\linewidth}
  \renewcommand\arraystretch{1.3}
  \Huge
   \begin{tabular}{|l|llll|l|}
\hline
\multicolumn{1}{|c|}{\multirow{2}{*}{\textbf{Reference}}} & \multicolumn{4}{c|}{\textbf{Details}}                                                                                                                                                                   & \multicolumn{1}{c|}{\multirow{2}{*}{\textbf{Reported Performance}}} \\ \cline{2-5}
\multicolumn{1}{|c|}{}                                    & \multicolumn{1}{c|}{\textbf{Task}}  & \multicolumn{1}{c|}{\textbf{Dataset}}                                                       & \multicolumn{1}{c|}{\textbf{Language}}  & \multicolumn{1}{c|}{\textbf{Domain}}                            & \multicolumn{1}{c|}{}                                               \\ \hline

 {\citet{fu2020facilitating}}    & \multicolumn{1}{l|}{\begin{tabular}[c]{@{}l@{}}Style Transfer\\(Paraphrasing)\end{tabular}}  & \multicolumn{1}{l|}  {\begin{tabular}[c]{@{}l@{}}Stanford Politeness Corpus\\of requests\end{tabular}}     &  \multicolumn{1}{l|} {\begin{tabular}[c]{@{}l@{}}English\end{tabular}}        & \begin{tabular}[c]{@{}l@{}} Online community\\( Wikipedia) \end{tabular}                                                                                                                                   & BLEU-s: 0.688                                                                                                                                                                    \\ \hline

 {\citet{madaan2020politeness}}    & \multicolumn{1}{l|}{\begin{tabular}[c]{@{}l@{}}Style Transfer\end{tabular}}  & \multicolumn{1}{l|}  {\begin{tabular}[c]{@{}l@{}}Enron Corpus\end{tabular}}     &  \multicolumn{1}{l|} {\begin{tabular}[c]{@{}l@{}}English\end{tabular}}        & \begin{tabular}[c]{@{}l@{}} Emails\end{tabular}                                                                                                                                   & BLEU-s: 0.704                                                                                                                                                                    \\ \hline

 {\citet{sennrich2016controlling}}  & \multicolumn{1}{l|} {\begin{tabular}[c]{@{}l@{}}NMT\end{tabular}}      & \multicolumn{1}{l|}  {\begin{tabular}[c]{@{}l@{}}OpenSubtitles\end{tabular}}      &  \multicolumn{1}{l|} {\begin{tabular}[c]{@{}l@{}}English$\rightarrow$German\end{tabular}}         & Movie subtitles                                                                                                                                       & BLEU: 0.24                                                                                                                                                                       \\ \hline
 {\citet{feely2019controlling}}   & \multicolumn{1}{l|} {\begin{tabular}[c]{@{}l@{}}NMT\end{tabular}}      & \multicolumn{1}{l|}  {\begin{tabular}[c]{@{}l@{}}ASPEC, JESC,\\KFTT\end{tabular}}      &  \multicolumn{1}{l|} {\begin{tabular}[c]{@{}l@{}}English$\rightarrow$Japanese\end{tabular}}                      & \begin{tabular}[c]{@{}l@{}}ASPEC (Scientific paper abstracts corpus), \\ JESC (movie and television subtitles), \\ KFTT (Wikipedia data)\end{tabular} & \begin{tabular}[c]{@{}l@{}}BLEU: 0.433 (ASPEC), 0.203 (JESC), \\ 0.255 (KFTT)\end{tabular}                                                                                          \\ \hline
 {\citet{niu2018polite}}   & \multicolumn{1}{l|} {\begin{tabular}[c]{@{}l@{}}Style Transfer\end{tabular}}      & \multicolumn{1}{l|}  {\begin{tabular}[c]{@{}l@{}}MovieTriples\end{tabular}}      &  \multicolumn{1}{l|}  {\begin{tabular}[c]{@{}l@{}}English\end{tabular}}                              & \begin{tabular}[c]{@{}l@{}}IMSDB movie scripts\end{tabular}                                                & BLEU-4: 0.94, Politeness Score: 61.0                                                                                                                                             \\ \hline
 {\citet{golchha2019courteously}}   & \multicolumn{1}{l|} {\begin{tabular}[c]{@{}l@{}}Dialogue\\Generation\end{tabular}}      & \multicolumn{1}{l|}  {\begin{tabular}[c]{@{}l@{}}CYCCD\end{tabular}}      &  \multicolumn{1}{l|}  {\begin{tabular}[c]{@{}l@{}}English\end{tabular}}                     & Customer care                                                                                                                           & \begin{tabular}[c]{@{}l@{}}BLEU: 0.692, ROUGE-1: 73.5, ROUGE-2: 69.9\\ ROUGE-L: 72.3, PPL: 0.437, CP: 77.5, EA: 86.8\end{tabular}                                                  \\ \hline

 {\citet{firdaus2020incorporating}} & \multicolumn{1}{l|} {\begin{tabular}[c]{@{}l@{}}Dialogue\\Generation\end{tabular}}      & \multicolumn{1}{l|}  {\begin{tabular}[c]{@{}l@{}}CYCCD\\Hindi Customer Care Dialogue\end{tabular}}      &  \multicolumn{1}{l|} {\begin{tabular}[c]{@{}l@{}}English\\Hindi\end{tabular}}                     & Customer care                                                                                                               & \begin{tabular}[c]{@{}l@{}}English:\\ BLEU: 0.724, ROUGE-L: 75.2, PPL: 0.418, CP: 79.2, EA: 87.9 \\ Hindi:\\ BLEU: 0.596, ROUGE-L: 61.4, PPL: 0.441, CP: 68.5, EA: 77.9\end{tabular} \\ \hline

 {\citet{wang2020can}}    & \multicolumn{1}{l|} {\begin{tabular}[c]{@{}l@{}}Dialogue\\Generation\end{tabular}}      & \multicolumn{1}{l|} {\begin{tabular}[c]{@{}l@{}}CSR-Driver Exchanges\end{tabular}}      &  \multicolumn{1}{l|} {\begin{tabular}[c]{@{}l@{}}English\end{tabular}}                    & \begin{tabular}[c]{@{}l@{}} Customer care\end{tabular}                                                                                                        & \begin{tabular}[c]{@{}l@{}}BLEU: 9.89, word2vec-based similarity: 9.38\end{tabular}                                                  \\ \hline

 {\citet{firdaus2022being}}   & \multicolumn{1}{l|} {\begin{tabular}[c]{@{}l@{}}Dialogue\\Generation\end{tabular}}      & \multicolumn{1}{l|} {\begin{tabular}[c]{@{}l@{}}bAbI\end{tabular}}      &  \multicolumn{1}{l|}  {\begin{tabular}[c]{@{}l@{}}English\end{tabular}}                   &  \begin{tabular}[c]{@{}l@{}}Personalized conversations\\ with polite templates \end{tabular}                                                                                      & \begin{tabular}[c]{@{}l@{}}BLEU-4: 0.275, ROUGE-L: 35.2, PPL: 1.004, PA: 73.0\end{tabular} \\ \hline

 {\citet{mishra2022please}}     & \multicolumn{1}{l|} {\begin{tabular}[c]{@{}l@{}}Dialogue\\Generation\end{tabular}}      & \multicolumn{1}{l|} {\begin{tabular}[c]{@{}l@{}}Politeness-annotated MultiDoGo\end{tabular}}     &  \multicolumn{1}{l|} {\begin{tabular}[c]{@{}l@{}}English\end{tabular}}                         & \begin{tabular}[c]{@{}l@{}}Task-oriented conversations \\ covering different domains\end{tabular}                                                     & \begin{tabular}[c]{@{}l@{}}Success rate: 65.8 (airline), 78.2 (fastfood), 71.4 (finance), \\ 72.4 (insurance), 73.2 (media), 63.2 (software)\\ PA: 47.0, TCR: 53.0\end{tabular}  \\ \hline
 {\citet{silva2022polite}}    & \multicolumn{1}{l|} {\begin{tabular}[c]{@{}l@{}}Polite\\Response\\Generation\end{tabular}}      & \multicolumn{1}{l|} {\begin{tabular}[c]{@{}l@{}}Enron and\\ fashion domain data \end{tabular}}      &  \multicolumn{1}{l|} {\begin{tabular}[c]{@{}l@{}}English\end{tabular}}                           & \begin{tabular}[c]{@{}l@{}}Emails\\ and fashion\end{tabular}                             & \begin{tabular}[c]{@{}l@{}}BLEU: 0.703, ROUGE: 86.7, METEOR: 51.5  {(Enron)}\\
BLEU: 0.850, ROUGE: 83.7, METEOR: 58.99  {(Fashion)}
%Politeness: 2.49, Grammar: 79.0
\end{tabular}                                                                   \\ \hline
 {\citet{saha2022countergedi}} & \multicolumn{1}{l|} {\begin{tabular}[c]{@{}l@{}}Counterspeech\\suggestions\end{tabular}}      & \multicolumn{1}{l|}  {\begin{tabular}[c]{@{}l@{}} CONAN\\Reddit\\Gab\end{tabular}}      &  \multicolumn{1}{l|} {\begin{tabular}[c]{@{}l@{}}English\end{tabular}}                             & \begin{tabular}[c]{@{}l@{}}Hatespeech instances \end{tabular}                             & \begin{tabular}[c]{@{}l@{}} BLEU-2: 0.136 (CONAN), 0.083 (Reddit), 0.088 (Gab) \\ COLA: 79 (CONAN), 81 (Reddit), 85 (Gab)%METEOR: 18 (CONAN), 17 (Reddit), 17 (Gab)\\ %Politeness: 2.49, Grammar: 79.0
\end{tabular}                                                                   \\ \hline

 {\citet{mishra2023help}} & \multicolumn{1}{l|} {\begin{tabular}[c]{@{}l@{}}Dialogue\\Generation\end{tabular}}      & \multicolumn{1}{l|}  {\begin{tabular}[c]{@{}l@{}}MHLCD\end{tabular}}      &  \multicolumn{1}{l|} {\begin{tabular}[c]{@{}l@{}}English\end{tabular}}                             & \begin{tabular}[c]{@{}l@{}}Mental health and\\ legal counseling\end{tabular}                             & \begin{tabular}[c]{@{}l@{}} CoStr: 80.30, Pol: 92.54, Emp: 46.4,\\ PPL: 1.91, R-LEN: 18.71
\end{tabular}                                                                   \\ \hline

 {\citet{mishra2023pal}} & \multicolumn{1}{l|} {\begin{tabular}[c]{@{}l@{}}Dialogue\\Generation\end{tabular}}      & \multicolumn{1}{l|}  {\begin{tabular}[c]{@{}l@{}}EPE-HLCC,\\EPE-enEIH\end{tabular}}      &  \multicolumn{1}{l|} {\begin{tabular}[c]{@{}l@{}}English\end{tabular}}                             & \begin{tabular}[c]{@{}l@{}}Mental health and/or\\ legal counseling\end{tabular}                             & \begin{tabular}[c]{@{}l@{}} EPC: 61.4, EEC: 60.7, PC: 65.8, EC: 63.1,\\PPL: 12.19, R-LEN: 25.81  {(EPE-HLCC)}\\
EPC: 73.7, EEC: 75.9, PC:71.6, EC: 75.2,\\ PPL: 2.03, R-LEN: 21.09  {(EPE-enEIH)}
\end{tabular}                                                                   \\ \hline

 {\citet{ijcai2023p686}} & \multicolumn{1}{l|} {\begin{tabular}[c]{@{}l@{}}Dialogue\\Generation\end{tabular}}      & \multicolumn{1}{l|}  {\begin{tabular}[c]{@{}l@{}} HEAL\end{tabular}}      &  \multicolumn{1}{l|} {\begin{tabular}[c]{@{}l@{}}English\end{tabular}}                             & \begin{tabular}[c]{@{}l@{}}Mental health and\\ legal counseling\end{tabular}                             & \begin{tabular}[c]{@{}l@{}} CoAct: 56.5, EmpStr: 61.8, PolStr: 69.9,\\ PPL: 2.55, R-LEN: 16.06
\end{tabular}                                                                   \\ \hline

 {\citet{mishra2023genpads}} & \multicolumn{1}{l|} {\begin{tabular}[c]{@{}l@{}}Dialogue\\Generation\end{tabular}}      & \multicolumn{1}{l|}  {\begin{tabular}[c]{@{}l@{}} Politeness-annotated Taskmaster\end{tabular}}      &  \multicolumn{1}{l|} {\begin{tabular}[c]{@{}l@{}}English\end{tabular}}                             & \begin{tabular}[c]{@{}l@{}}Restaurants, food ordering,\\movie, \\flight,\\hotels, music\\sports\end{tabular}                             & \begin{tabular}[c]{@{}l@{}}DL:10.7, POL:85.1, MET:72.1, R-2-F1:54.3, SR:79  {(Flights)}\\
DL:11.5, POL:93.6, MET:64.2, R-2-F1:44.4, SR:86  {(Food)}\\
DL:9.9 11.5, POL:89.3, MET:70.9, R-2-F1:57.3, SR:82  {(Hotels)}\\
DL:9.5, POL:88.8, MET:70.0, R-2-F1:40.7, SR:84  {(Movies)} \\
DL:9.4, POL:95.9, MET:44.0, R-2-F1:23.7, SR:86  {(Music)} \\
DL:8.5, POL:92.0, MET:70.9, R-2-F1:46.7, SR:82  {(Restaurant)}\\ 
DL:10.9, POL:94.8, MET:65.7, R-2-F1:41.8, SR:81  {(Sports)}\\
\end{tabular}  
\\ \hline

 {\citet{firdaus2022polise}} & \multicolumn{1}{l|} {\begin{tabular}[c]{@{}l@{}}Dialogue\\Generation\end{tabular}}      & \multicolumn{1}{l|}  {\begin{tabular}[c]{@{}l@{}} Sentiment-annotated CYCCD\end{tabular}}      &  \multicolumn{1}{l|} {\begin{tabular}[c]{@{}l@{}}English\end{tabular}}                             & \begin{tabular}[c]{@{}l@{}}Customer care\end{tabular}                             & \begin{tabular}[c]{@{}l@{}}PPL: 1.004, BLEU-4:  0.275,\\ ROUGE-L: 0.352, PA:0.77
\end{tabular}                                                                   \\ \hline

 {\citet{mishra2022pepds}} & \multicolumn{1}{l|} {\begin{tabular}[c]{@{}l@{}}Dialogue\\Generation\end{tabular}}      & \multicolumn{1}{l|}  {\begin{tabular}[c]{@{}l@{}} Politeness-annotated \\ PersuasionForGood\end{tabular}}      &  \multicolumn{1}{l|} {\begin{tabular}[c]{@{}l@{}}English\end{tabular}}                             & \begin{tabular}[c]{@{}l@{}}Charity\end{tabular}                             & \begin{tabular}[c]{@{}l@{}}  PerStr: 59.98, PolSt: 41.11,  Emp: 78.1,\\ PPL: 11.06, LEN: 16.87
\end{tabular}                                                                   \\ \hline

 {\citet{firdaus2023mixing}} & \multicolumn{1}{l|} {\begin{tabular}[c]{@{}l@{}}Dialogue\\Generation\end{tabular}}      & \multicolumn{1}{l|}  {\begin{tabular}[c]{@{}l@{}} Polite Behavior-annotated \\ CYCCD\end{tabular}}      &  \multicolumn{1}{l|} {\begin{tabular}[c]{@{}l@{}}English\end{tabular}}                             & \begin{tabular}[c]{@{}l@{}}Customer Care\end{tabular}                             & \begin{tabular}[c]{@{}l@{}} PPL: 0.987, BLEU-4: 0.310, \\ ROUGE-L: 38.2, PA: 81  
\end{tabular}                                                                   \\ \hline

 {\citet{mishra2023therapist}} & \multicolumn{1}{l|} {\begin{tabular}[c]{@{}l@{}}Dialogue\\Generation\end{tabular}}      & \multicolumn{1}{l|}  {\begin{tabular}[c]{@{}l@{}} PSYCON\end{tabular}}      &  \multicolumn{1}{l|} {\begin{tabular}[c]{@{}l@{}}English\end{tabular}}                             & \begin{tabular}[c]{@{}l@{}}Mental health counseling \end{tabular}                             & \begin{tabular}[c]{@{}l@{}}
GA$_c$: 90.1, P$_c$: 84.1, CT$_c$: 92.6, Po$_c$: 92.5, IB$_c$: 83.4 \\
PPL: 2.52, BERTScore-F1: 0.89, R-LEN: 23.89
\end{tabular}                                                                   \\ \hline

\end{tabular}
\end{adjustbox}
\end{table*}

%%%%%%%%%%%%%%%%%%%%%%%%%%%%%%%%%%%%%%%%%%%%%%%%%%%%%%%%%%%%%%%%%%%%%%%%%%%%%%%%

\textbf{Reported Performance on Politeness Generation.} There have been several works that attempt to incorporate politeness in natural language generation systems. Table \ref{tab:results_generate} depicts the performance evaluation for such systems as documented in the existing literature. This evaluation encompasses various datasets across diverse domains. For instance, \citet{madaan2020politeness} uses the n-grams tf-idf-based approach to estimate the politeness markers and generate the politeness-oriented text. They have achieved a BLEU score of 0.704. Likewise, \citet{fu2020facilitating} achieves a BLEU \cite{papineni2002bleu} score of 0.688 for generating the paraphrased version of a given message that can convey the desired level of politeness. \citet{niu2018polite} incorporate politeness in chit-chat dialogues and achieve a BLEU score of 0.24. 

Lately, \citet{golchha2019courteously,firdaus2020incorporating} attempt to induce politeness in task-oriented dialogues. They have used both automatic evaluation (Perplexity (PPL) \cite{brown1992estimate}, Emotion Accuracy (EA), Content Preservation (CP), BLEU, ROUGE \cite{lin2004rouge}) as well as human evaluation (Fluency, Content Adequacy, and Politeness Consistency) metrics to evaluate the performance of their proposed framework. \citet{mishra2022please} utilize task success rate and human evaluation metrics, namely Task Completion Rate (TCR) and Politeness Adaptability (PA), to show the effectiveness of their proposed framework. \citet{saha2022countergedi} generate polite, detoxified, and emotionally charged counter-speech using three hate-speech datasets, namely CONAN \cite{chung2019conan}, Reddit \cite{qian2019benchmark}, and Gab \cite{qian2019benchmark}, and observe significant improvement across different attribute scores. The politeness and detoxification scores increased by around 15\% and 6\%, respectively, while the emotion in the counter-speech increased by at least 10 points across all the datasets. 

In recent times, there have been works that employ task-specific metrics to measure the level of politeness and other related aspects like emotion and empathy in the generated responses. For instance, in \citet{firdaus2022being,firdaus2022polise}, the authors utilize Politeness Accuracy (PA) to measure the degree of politeness in the generated responses. \citet{mishra2023help}, \citet{mishra2023pal}, \citet{ijcai2023p686} and  \citet{mishra2022pepds} employ several task-specific metrics, \textit{viz.} <CoStr (Counseling Strategy Consistency), Pol (percentage of polite utterances generated), Emp (percentage of empathetic utterances generated)>, <PC (politeness correctness), EC(Emotion Correctness), EPC (Emotion-Politeness Consistency), EEC (Emotion-Empathy Consistency)>, <PolStr (Politeness Strategy Consistency), EmpStr (Empathy Strategy Consistency), CoAct (Counseling Act Correctness)>, and <PolSt (Politness Strategy Consistency), Emp (Empathy), PerStr (Persuasion Strategy Consistency)>, respectively to evaluate the efficacy of their approach in terms of task success. It is to be noted that these works employ generic metrics like PPL (Perplexity) and R-LEN/DL (Response Length/Dialogue Length) as well for evaluating the generated responses.

Most of the past works do not provide an analysis of systems based on the variation of politeness across different domains, demographics, etc., mainly owing to the absence of corpora with varying degrees of politeness specified. However, an analysis of politeness transfer across domains is seen in \citet{silva2022polite}. They look at the examples that their model correctly generates polite responses through rewriting and conclude that politeness through rewriting provides a balance between the delivery of polite expressions and the preservation of contents. Recently, \citet{firdaus2022being} analyze politeness variation across different age groups and genders, reporting a Politeness Appropriateness (PA) of 73\% as a measure of the politeness quotient of response. Subsequently, \citet{mishra2023therapist} explore variations in politeness and interpersonal behavior among individuals of different age groups, genders, and personas. Their analysis involved assessing Gender-Age Consistency (GA$_c$), Persona Consistency (P$_c$), Psychotherapeutic Approach Correctness (CT$_c$), Politeness Correctness (Po$_c$), and Interpersonal Behavior Correctness (IB$_c$), yielding values of 90.1\%, 84.1\%, 92.6\%, 92.5\%, and 83.4\%, respectively.

%%%%%%%%%%%%%%%%%%%%%%%%%%%%%%%%%%%%%%%%%%%%%%%%%%%%%%%%%%%%%%%%%%%%%%%%%%%%%%%%%%%%%%%%%%%%%%%%%%%%%
\begin{figure}
    \centering
    \begin{adjustbox}{max width=\linewidth}
    \includegraphics[scale=1.0]{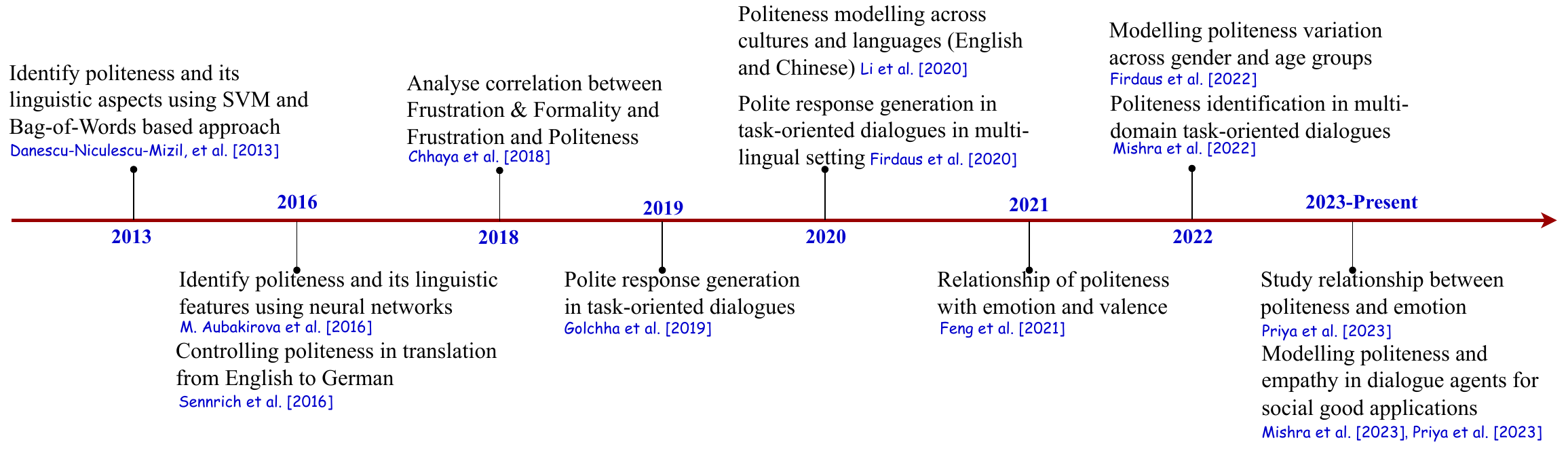}
    \end{adjustbox}
    \caption{Trends in computational politeness research}
    \label{trends}
    \vspace{-14pt}
\end{figure}
%%%%%%%%%%%%%%%%%%%%%%%%%%%%%%%%%%%%%%%%%%%%%%%%%%%%%%%%%%%%%%%%%%%%%%%%%%%%%%%%%%%%%%%%%%%%%%%%%%%%%

%==================================================================================================%

\section{Trends in Computational Politeness}
\label{section7}
In the preceding sections of this paper, we examined the datasets, approaches, and performance values of past works in computational politeness studies in Natural Language Processing. In this section, we delve into trends observed in computational politeness research. These trends are depicted in Figure \ref{trends}. %Representative studies in each area are shown in the figure, around four key milestones. At first, 
In earlier research works, the supervised approaches for politeness detection were investigated. These methods concentrated on utilizing specific features for identifying politeness in natural language text. Following fundamental studies, recently, there has been a growing trend to employ context beyond the target text for comprehending and generating polite text. Furthermore, there is a growing interest in the computational approach for modeling the link between politeness and other socio-linguistic cues; the variations in politeness across different cultures, age groups, genders, and social power; and the study of politeness for various social good applications.

%\subsection{Use of Context in Computational Politeness Research}

\textbf{Use of Context in Computational Politeness Research.} Using context beyond the text to be predicted is becoming a popular trend in computational politeness research. The term context here refers to any information outside the text to be predicted and outside the common knowledge. For example, the sentence \textit{\enquote{Let's have a drink}} may not be polite for a new person in the group but may be polite to many others. This type of situation needs knowledge beyond the text to be categorized. In the following part of this section, we refer to the textual unit to be classified as the \enquote{target text}. 

The potential of context for politeness study has sparked the interest of researchers to look at the approaches for incorporating it. The contextual information may be of various types depending upon the problem definition; for instance, in automatic sarcasm detection, the context may come from the authors' information, topical information, or conversations of which the sarcastic utterance is a part. In computational politeness research, incorporating the conversational context is prevalent for identifying and generating politeness. By conversational context, we mean the conversational text to which the target text belongs. Consider a simple exclamation, \textit{\enquote{Of course!}}. This exclamation may or may not be polite. To understand the politeness in the given utterance, it is necessary to look at the conversation to which the target text belongs. If the utterance %preceding 
before the exclamation is \textit{\enquote{Do you want me to help you with your problem?}}, then the impoliteness in the exclamation can be inferred. On the contrary, if the utterance preceding the exclamation is \enquote{\textit{Can I have your booking preference?}}, then the utterance, \enquote{\textit{Of course!}} here implies \enquote{\textit{Yes! obviously, I’m happy to share the preferences with you.}}, and is thus interpreted as polite.

\citet{mishra2022predicting} develop a computational framework for identifying %and characterizing 
politeness in goal-oriented dialogue systems considering the context from previous utterances. They have utilized the hierarchical transformer network to obtain the context-sensitive representation of the utterances. \citet{golchha2019courteously} utilize the contextual information from previous utterances for transforming neutral customer care responses into polite ones. \citet{firdaus2020incorporating} extend the work of \citet{golchha2019courteously} and generate polite responses in both English and Hindi while being contextually coherent with the previous conversation turns. \citet{mishra2022please} use numerous politeness-related para-linguistic features \cite{brown1978universals} and build %an end-to-end 
a conversational agent with a polite-to-impolite spectrum based on a politeness classifier without relying on parallel data. The authors consider conversational history as the context for politeness classification. 

% \subsection{Modeling Relationship between Politeness and Social Factors}

\textbf{Modeling Relationship between Politeness and Social Factors. }Due to the importance of politeness in cross-cultural scenarios, there is a recent trend to model politeness across cultures and languages. \citet{li2020studying} study the differences and similarities in politeness expressions between American English and Mandarin Chinese. The authors have developed machine learning algorithms for predicting politeness in both languages. \citet{sperlich2016interaction} examine the system of politeness covering French \textit{tu} and \textit{vous} from the perspective of a Korean learner of French. They state that the acquisition of \textit{tu} is comparatively easier due to positive transfer than \textit{vous}'s acquisition because it involves negative transfer. The explicit/implicit state of the learners’ pragmalinguistic and sociopragmatic knowledge is another factor affecting the use and interpretation of \textit{tu/vous}. 

Natural language is the most powerful tool for performing speech acts in human communication (such as making requests, giving commands, apologizing, expressing gratitude, etc.). These speech acts are executed in accordance with specific norms and principles. One of these principles is politeness. \citet{munkova2013identifying} propose a novel approach to computational modeling of specific features of politeness in the speech act of requests in written forms. It examines the similarities and differences in using specific social and expressive factors in the primary language (L1) and the foreign language (L2, English). In particular, the authors formulate the Transaction/Sequence model for text representation, apply the association rules for analysis, and indicate that the requests written in the primary language are less direct than those written in a foreign language. A few studies explore the link between politeness and social variables, focusing on social power dynamics. The work of \citet{danescu2013computational} is the earliest attempt in this direction. The authors have used their computational framework to study the relationship between politeness and power. They have shown that polite editors are likely to gain higher status through elections; however, once they are at the top of their reputation, their politeness decreases. They have also demonstrated the utility of their framework in exploring the differences in politeness across other factors of interest, such as communities, geographical regions, and gender. The authors reveal that (i). different programming communities vary significantly by politeness on Stack Exchange; (ii). Wikipedia editors from the U.S. Midwest are most polite, and (iii). female editors on Wikipedia are generally more polite compared to their male counterparts.

Some studies evaluate the effect of people's age and gender on human-machine interaction and discuss politeness as an essential social skill to develop trust in such interactions and to ensure optimal interactions between machines and humans. \citet{inbar2019politeness} conduct tests to examine how people behave when a more or less courteous robot responsible for access control interacts with individuals of varying ages and genders. Their study reveals that politeness is the only factor that influences the participant's perception of the peacekeeping robot's behavior. Polite peacekeeping robots are perceived as friendlier, fairer, less intimidating, and more appropriate. However, they further demonstrate that the age and gender of the people interacting with the robots had no significant effect on participants’ impressions of its attributes. The authors claim that their findings are robust as the studies involved male and female participants of a wide range of ages. They have explained in support of these findings that when the gap between a polite and an impolite robot’s behavior is so pronounced, it overshadows other distinctions of age and gender. Latterly, \citet{firdaus2022being} study politeness variations across different age groups and genders. They utilize the age and gender of the user to decide the degree of politeness in a response. The authors state that system responses for an \textit{elderly} person should be more polite as compared to the responses for a \textit{young} person of the same gender. Likewise, the response for the \textit{elderly female} user has a slightly higher politeness quotient than the \textit{elderly male} user. 

%\subsection{Modeling Relationship between Politeness and Other Socio-linguistic Cues}
\textbf{Modeling Relationship between Politeness and Other Socio-linguistic Cues. }Lately, research has been inclined towards exploring the relationship between politeness and other socio-linguistic cues in interaction. Besides politeness, many socio-linguistic cues are used in the conversational analysis, such as emotion, sentiment, and dialogue acts. Detecting politeness in communication offers information about the interlocutor's social demeanor. Likewise, perceiving emotions provides information about the participants' emotional states. Sentiment or valence is another dimension in conversational analysis that helps detect the positivity and negativity of emotion, and identifying dialogue acts gives information about the performative function of an utterance. \citet{bothe2021polite} attempt to correlate the relational bonds between politeness and emotions, and politeness and dialogue acts. In particular, the authors analyze the utterances in the  DailyDialog dataset \cite{li2017dailydialog} against the emotion and dialogue act classes in the dataset. The authors have discovered that most of the utterances in \textit{Sadness} and \textit{Happiness} emotion classes are polite, and the utterances with emotion classes, \textit{Anger} and \textit{Disgust}, are more likely to tend to be impolite. A similar phenomenon occurs with dialogue acts; \textit{Inform} and \textit{Commissive} contain more polite utterances than \textit{Question} and \textit{Directive} dialogue acts. 

\citet{feng2021emowoz} use the Ortony, Clore, and Collins (OCC) model \cite{ortony1988cognitive} and consider three aspects, \textit{viz.} Elicitor or Cause, Valence, and Conduct for annotating the utterances in the dataset with the appropriate emotion classes. The
authors mention that in conversational analysis, it is essential to distinguish between user and system utterances and identify valence as it is highly related to task success or failure, thus expressing the positivity or negativity of emotion. Successfully achieving a goal is likely to evoke positive valence, while a misunderstanding with an agent or failure to achieve the objective will likely cause negative valence. Moreover, conduct that describes the politeness of users can indicate the degree of valence. For instance, the user can express extreme dissatisfaction through rudeness. It also helps to differentiate between emotions, such as those linked with apology or anger, both of which are intrinsically negative. The apologetic emotion is more likely to be associated with the polite demeanor of the interlocutor, and anger, by contrast, is more likely to be associated with impolite behavior. The authors have eventually created a set of seven emotions that differ in terms of elicitor, valence, and conduct to reflect the cognitive context of emotions while preserving the simplicity of labeling.  

Following this line of the work, \citet{priya2023multi} investigate the inter-connectedness between politeness and emotion and devise a novel multi-task learning (MTL) framework, Caps-DGCN for simultaneous Politeness and Emotion Detection (PED) in conversations. The authors employ the Capsule Network (Caps) \cite{du2019investigating} to capture the local ordering of words in the utterance and corresponding semantic representations, and the Directional Graph Convolutional Network (DGCN) \cite{chen2020joint} to encode the syntactic information by incorporating the dependency among the words. The authors mention that the identification of politeness and emotion in conversations requires the extraction of semantic as well as syntactic information from an utterance. The authors established that their proposed Caps-DGCN framework strengthens the utterance representation by utilizing semantic and syntactic information and effectively captures the relatedness between politeness and emotion in utterances. 

%\subsection{Modeling Politeness in Dialogue Systems for Social Good Applications}
\textbf{Modeling Politeness in Dialogue Systems for Social Good Applications. }Recently, there have been studies that investigate the role of politeness and its complementary cues (emotion, empathy, etc.) in dialogue systems for various social good applications like persuasion \cite{mishra2022pepds}, mental health support \cite{mishra2023help,mishra2023pal,ijcai2023p686}, to mention a few. \citet{mishra2022pepds} devise a polite and empathetic dialogue system for persuasion tasks in the charity domain. The authors hypothesize that incorporating polite tone and emotional quotient into persuasive dialogue systems may make persuasive messages more appealing to the user, resulting in better persuasive outcomes. To develop a polite and empathetic persuasive dialogue system (PEPDS), the authors have fine-tuned a Maximum Likelihood Estimation loss-based dialogue model in an RL setting by designing a novel reward function. This reward function comprises five sub-rewards, \textit{viz.} Politeness-strategy consistency, Persuasion, Emotion, Dialogue-Coherence, and Non-repetitiveness to ensure politeness-strategy consistency, persuasiveness, emotion acknowledgment, dialogue-coherence, and non-repetitiveness, respectively, in the generated responses. Afterward, the authors build an Empathetic transformer model on top of the RL fine-tuned model to transform non-empathetic utterances into empathetic ones. 

The authors in  \cite{mishra2023help,mishra2023pal,ijcai2023p686, mishra2023therapist} propose counseling dialogue systems capable of conversing with the users politely. \citet{mishra2023help} propose a polite and empathetic mental health and legal counseling dialogue system (\textbf{Po-Em-MHLCDS}) for crime victims. To develop \textbf{Po-Em-MHLCDS}, the authors have alternatively trained two different GPT2-medium \cite{radford2019language} based language models to learn the distribution of the victim's and counselor's utterances to achieve natural language interaction between both the models. This MLE loss-based agent's trained model is then fine-tuned in an RL framework. In order to optimize the agent's behavior, the authors employ an RL loss considering an efficiently designed reward function to ensure appropriate counseling strategy, politeness, empathy, contextual coherence, and non-repetitiveness while counseling the victim. 

\citet{mishra2023pal} introduce an emotion-adaptive polite and empathetic counseling conversational agent \textbf{PAL}. To develop \textbf{PAL} that can generate polite and empathetic responses while reflecting the client's emotional state naturally and without unnecessary repetition, the authors have trained a BART-large \cite{lewis2019bart} based client's emotion-informed dialogue model (\textit{EIDM}) in a supervised learning framework. The \textit{EIDM} is then fine-tuned with a Proximal Policy Optimization (PPO) \cite{schulman2017proximal} loss utilizing novel designed reward function consisting of preference rewards, task-specific rewards, and generic rewards. \citet{ijcai2023p686} present a polite and empathy strategies-adaptive mental health and legal counseling dialogue system, named \textbf{PARTNER} that can adopt appropriate politeness and empathy strategy together with the pertinent counseling act, including but not limited to persuasion during the counseling sessions. The authors mention that \textbf{PARTNER} learns from user interactions and improves itself depending on user feedback through rewards. In particular, the authors have initially trained a Cross-Entropy Loss based Dialogue Model (\textit{CELDM}) - to foster natural language interaction between the \textit{counseling bot} and the \textit{crime victim}, and then fine-tuned the trained \textit{CELDM} using Proximal Policy Optimization (PPO) \cite{schulman2017proximal} Loss based Dialogue Model (\textit{PLDM}) - to generate politeness and empathy strategies-adaptive counseling responses. The GPT-small \cite{radford2019language} is used as the backbone architecture for \textit{CELDM} model.

\citet{mishra2023therapist} propose e-Therapist, a polite and interpersonal dialogue system for psychotherapy that generates responses according to the user’s age, gender, persona, and sentiment. e-Therapist is developed by first fine-tuning a language model using a supervised learning approach, which is referred to as SLLM. The SLLM then generates a set of candidate responses which are ranked using a novel reward model employing gender-age awareness, persona awareness, counseling, politeness, and interpersonal behavior rewards to reinforce the counseling agent to ensure the preferences of gender, age, persona, and sentiment-aware polite interpersonal psychotherapeutic responses. This reward model is finally optimized using the NLPO loss \cite{Ramamurthy2022IsRL} to yield the counseling dialogue system, e-Therapist.
%==================================================================================================%

\section{Issues in Computational Politeness}
\label{section8}

In this section, we discuss the critical issues that appear in different computational politeness works. The most important issue that is faced in most computational politeness research is the lack of annotated data. 

With the success of machine learning and deep learning-based architectures, there has been a surge in interest in utilizing such architectures for computational politeness studies. It must, however, be noted that these approaches are data-intensive, and publicly available data are scarce for computational politeness study. \citet{danescu2013computational} propose a dataset of requests annotated for politeness, which may be considered ground-breaking work in this direction. Following this work, many research works such as  \citet{aubakirova2016interpreting,li2020studying} use this corpus to train classifiers for identifying politeness. \citet{mishra2022please,mishra2022predicting} also utilize the classifier used in \citet{danescu2013computational} for creating the conversational dataset for their work. \citet{niu2018polite,golchha2019courteously,madaan2020politeness,wang2020can} are the few other works that introduce corpus for computational modeling of politeness. %There have been 
A few corpora have been available for modeling politeness in languages other than English. For instance,  \citet{firdaus2020incorporating,kumar2021towards} introduce dataset in Hindi, and \citet{li2020studying} introduce Chinese corpora annotated for politeness. Many of these datasets are created manually, either through crowdsourcing or by the researchers themselves. Recently, within computational politeness research, there has also been a growing interest in using linguistic variation related to the identity of speakers, such as age, gender, demographics, etc., for improving the identification and generation of politeness in natural language texts. \citet{firdaus2022being} train gender and age-specific models for generating personalized and polite dialogue responses by manually creating the gender and age-specific polite templates. The manual transcription and annotation of data are time-intensive and costly.

Moreover, the quality of annotations is a significant concern in the case of manually annotated datasets. Because of the subjective nature of politeness, the inter-annotator agreement values mentioned in the previous research works are diverse. %diversified. 
\citet{li2020studying} report Krippendorff’s alpha \cite{krippendorff2007computing} inter-rater reliability of 0.528 for Twitter and 0.661 for Weibo. \citet{chhaya2018frustrated} indicate agreements for 3 class and 5 class politeness annotations as 0.64 ± 0.03. In \citet{golchha2019courteously,mishra2022predicting,mishra2022please}, the authors report a multi-rater Kappa \cite{mchugh2012interrater} agreement ratio of approximately 80\%. The value in the case of \citet{firdaus2020incorporating} is 90\%. This observation highlights the need to frame appropriate guidelines for annotators. \citet{danescu2013computational} present a corpus of requests annotated for politeness. They state that for politeness annotation, they asked the annotators to read a batch of requests, consider those requests as originating from a co-worker by email, and then annotate the requests with perceived politeness. \citet{li2020studying} provide the annotators with a list of politeness strategies along with their definitions and suitable examples from social media to annotate the Twitter and Weibo posts for perceived politeness values. \citet{golchha2019courteously,mishra2022please,mishra2022predicting} employ annotators who are highly proficient in English and have good exposure to the related task and explain them with suitable examples for politeness annotation.

%==================================================================================================%

\section{Conclusion and Future Directions}
\label{section9}

Over the past several years, research in computational approaches to politeness has received immense attention, prompting a retrospective evaluation of the overall picture that these individual works have produced. This article examines various approaches to computational politeness. We observed four milestones in the history of computational politeness research: supervised and weakly supervised feature extraction to identify and induce politeness in a given text, incorporation of context beyond the target text, a few studies of politeness across different social factors such as culture, age, power, gender, etc., and exploring the link between socio-linguistic indicators such as emotion, dialogue acts, etc. 

This survey paper furnished illustrations containing the datasets, approaches, and performance values as stated in previous works. The two popular techniques for constructing the politeness-annotated dataset include manual labeling for politeness and automatic labeling for politeness through supervision. We observed that statistical approaches utilize features like lexical, syntactic, affect-based, and so on. Various deep learning-based approaches to automatic politeness identification and generation have also been presented in the literature.  {Recently, we also witnessed a growing trend in evaluating the competency of LLMs in identifying, interpreting, and generating politeness.} Previously, additional conversation-specific features have been investigated to incorporate context. An underlying theme of several earlier methods is attempting to identify formality, positivity, and rapport building that is at the source of politeness. We further highlight the issues in computational politeness: lack of politeness-annotated corpora and quality of manual annotation. A comparative table is also provided, outlining key aspects of previous articles, including methodologies, annotation techniques, and features. We believe that the researchers who intend to comprehend the present state-of-the-art in computational politeness research will find this table informative and helpful.

Recent trends in computational politeness research emphasize contextual inclusion and modeling politeness in relation to various social and linguistic factors. While there are manually labeled gold-standard datasets, supervision is the prominent method for obtaining labels. Novel deep learning-based approaches have also been employed to infuse the context of different forms. Based on our examination of these works, we find the following emerging directions: 

%\begin{enumerate}

    %\item 
\textbf{Modeling politeness variation across different social factors. }A future area of research can be the automatic identification of variation in politeness across people of different age groups, genders, and cultures. A few approaches, such as \cite{danescu2013computational,li2020studying}, have explored such variation to a limited extent. Lately, \citet{firdaus2022being} and \citet{mishra2023therapist} generate polite responses based on the age and gender of the user. A thorough investigation and modeling of such variation would be helpful for the development of end-to-end human-like artificial intelligent systems. It facilitates understanding of the social attitude of the interlocutors, thereby fostering the interaction among them.    
    %\item 

\textbf{Modeling politeness variation across different socio-linguistic cues. }\citet{bothe2021polite} recently provided a corpus-based study and attempted to correlate the relational bonds between politeness and its complementary cues like emotion, dialog acts, etc.  {Following this, \citet{priya2023multi} employed a deep learning-based multi-task framework for exploring the association between politeness and emotion.} This work could be extended to learn social behaviors using analyzed linguistic features with the help of more advanced deep-learning techniques and LLMs. The relatedness of socio-linguistic variables also adds a new layer to the behavioral analysis of interlocutors and their use in the virtual assistant and human-robot interaction.
    
    %\item 
\textbf{Modeling of politeness for different domains and low-resource languages. }Politeness is perceived on the basis of shared knowledge. As indicated in past works such as  \citet{munkova2013identifying}, politeness is closely related to language-specific traits. The majority of the past works on identifying and incorporating politeness focus on the English language and a few domains, such as customer care and fashion. Future approaches to computational politeness will benefit from expanding politeness understanding for new languages like Hindi, Spanish, etc., and for new domains. 
    
%\item 
\textbf{Retrieval of contextual information using deep learning-based architectures. } Until now, only a handful of approaches have examined architectures based on deep learning. As mentioned in \cite{golchha2019courteously,firdaus2020incorporating,mishra2022please,mishra2022predicting}, extracting contextual information may be useful to acquire the additional shared knowledge (for example, conversation-specific or user-specific knowledge) that is required to understand politeness.

%\item  
{\textbf{Investigating politeness for social good applications. }In recent years, there has been a notable emergence in the exploration of politeness for various social good applications. From mental health counseling to persuasion techniques, researchers are delving deeper into understanding how politeness can positively influence various aspects of human interaction and behavior. A handful of studies in mental health counseling \cite{priya2023multi, mishra2023help, mishra2023pal, ijcai2023p686, mishra2023therapist} attempted to induce politeness in dialogue agents to foster trust, rapport, and overall therapeutic effectiveness between agents and users. Likewise, in the domain of persuasion, \cite{mishra2022pepds} incorporated politeness strategies to enhance the persuasiveness of messages and facilitate a more constructive and respectful atmosphere. Future approaches to computational politeness could explore politeness for other social good applications, like negotiation, education, environmental monitoring, etc., to build more empathetic, engaging, and effective communication channels that positively impact users' well-being and promote positive social change. } 
    
%\end{enumerate}

%==================================================================================================%

\begin{acks}
Priyanshu Priya acknowledges the Innovation in Science Pursuit for Inspired Research (INSPIRE) Fellowship implemented by the Department of Science and Technology, Ministry of Science and Technology, Government of India for financial support. Asif Ekbal acknowledges the Young Faculty Research Fellowship (YFRF), supported by Visvesvaraya PhD scheme for Electronics and IT, Ministry of Electronics and Information Technology (MeitY), Government of India, being implemented by Digital India Corporation (formerly Media Lab Asia).
\end{acks}

\bibliographystyle{ACM-Reference-Format}
\bibliography{main}

% \appendix









\end{document}